\documentclass{ieeetj}
\usepackage{cite}
\usepackage{amsmath,amssymb,amsfonts}
\usepackage{algorithmic}
\usepackage{gensymb} 
\usepackage{graphicx,color}
\usepackage{textcomp}
\usepackage{xcolor}
\usepackage{hyperref}
\hypersetup{hidelinks=true}
\usepackage{algorithm,algorithmic}
\def\BibTeX{{\rm B\kern-.05em{\sc i\kern-.025em b}\kern-.08em
    T\kern-.1667em\lower.7ex\hbox{E}\kern-.125emX}}
\AtBeginDocument{\definecolor{tmlcncolor}{cmyk}{0.93,0.59,0.15,0.02}\definecolor{NavyBlue}{RGB}{0,86,125}}

\usepackage{multirow}
\usepackage{rotating}
\usepackage[caption=false]{subfig}
\usepackage{siunitx}
\usepackage{amsmath,multirow,multicol}
\DeclareMathOperator{\sign}{sign}
\usepackage{cuted}
\usepackage{nccmath}
\usepackage{textgreek}
\usepackage{booktabs}
\usepackage{array}

\DeclareMathOperator*{\argmin}{argmin}

\def\OJlogo{\vspace{-4pt}}
\def\seclogo{\vspace{10pt}}

\def\authorrefmark#1{\ensuremath{^{\textbf{#1}}}}

\begin{document}


\title{Learning All-Terrain Locomotion for a Planetary Rover with Actively Articulated Suspension}

\author{Arthur Bouton\authorrefmark{1}, Tristan D. Hasseler\authorrefmark{1}, Michael Paton\authorrefmark{1}, Travis Brown\authorrefmark{1}, Jacob Levy\authorrefmark{2}, William Reid\authorrefmark{1}, Joshua Martin\authorrefmark{3}, Hari Nayar\authorrefmark{1}}
\affil{Jet Propulsion Laboratory, California Institute of Technology, Pasadena, CA 91109 USA}
\affil{Center for Autonomy, University of Texas at Austin, Austin, TX 78712 USA}
\affil{Space Systems Laboratory, University of Maryland, MD 20742 USA}
\corresp{Corresponding author: Arthur Bouton (email: arthur.bouton@jpl.nasa.gov).}
\authornote{The research described in this publication was carried out at the Jet Propulsion Laboratory, California Institute of Technology, under contract with the National Aeronautics and Space Administration.}

\begin{abstract}
    This paper presents ERNEST, a four-wheeled planetary rover concept equipped with a two-degree-of-freedom \emph{Active Gimbal Suspension} that combines yaw and roll actuation to enable wheel reconfiguration, steering, and active load redistribution. A single neural network controller, trained to track a desired path across challenging terrain, fully unlocks the capabilities of this actuated suspension system for autonomous obstacle negotiation. A reinforcement learning framework is developed using the high-fidelity DARTS simulation engine, which combines rigid-contact dynamics and Bekker--Wong terramechanics, enabling the emergence of locomotion strategies adapted to loose-soil conditions. To obtain a single unified controller across heterogeneous terrains, a policy consolidation strategy merges the experience of terrain-specialized agents into one neural network, eliminating the need for explicit terrain classification and controller switching. The resulting controller operates on a combination of proprioceptive and exteroceptive feedback, including sparse stereo-derived terrain elevation, chassis attitude, joint states, and force–torque measurements. Zero-shot transfer to the physical rover is achieved through domain randomization, sensor noise injection, and model-to-real system identification. Experimental results demonstrate autonomous traversal of rock fields, a Bickler trap (bump obstacle), a wheel-high step, sand ripples, and sandy slopes. On a \qty{20}{\degree} sandy slope, the learned controller reduces the cost of transport by \qty{37}{\percent} on dry sand despite the additional actuation, and achieves superior performance on wet sand where the passive suspension becomes completely immobilized. A video accompanying this paper is available at \url{https://youtu.be/d684P5a3xMc}.
\end{abstract}



\maketitle

\section{INTRODUCTION}

\subsection{OBJECTIVE}

Planetary rovers remain the most practical platform for long-range in situ surface exploration because wheeled locomotion offers energy efficiency, mechanical simplicity, and continuous ground-supported stability~\cite{fiorini2000ground,seeni2010robot,bruzzone2012locomotion,thoesen2021planetary}. However, the scientific objectives of future missions are expected to drive vehicles into increasingly demanding environments that challenge the limits of passive suspensions: steep granular slopes, sandy dune fields patterned with periodic ripples, and boulder-strewn rocky terrain~\cite{national2022origins}. Enabling traversal of such terrains would not only open access to new scientifically valuable sites, but also allow rovers to take shorter paths between targets. This, in turn, can increase mission productivity and may prove critical in scenarios where stopping in an unfavorable location risks mission loss, such as in permanently shadowed regions on the Moon or in the polar regions of Mars.

The objective of this work is to develop a system that can substantially expand the mobility envelope of a wheeled rover while preserving the simplicity and efficiency that make wheeled systems attractive for planetary missions. More specifically, this paper seeks to demonstrate that a four-wheeled rover equipped with a two-degree-of-freedom active suspension can autonomously traverse a variety of terrains intractable for passive suspensions, when governed by a sufficiently versatile controller.

\subsection{BACKGROUND}

Classical passive suspensions conform to terrain geometry under the effect of gravity. With an appropriate arrangement of passive joints, they can achieve near-uniform load distribution across the wheels around a nominal configuration~\cite{bickler1998roving}, and exhibit strong performance over specific classes of obstacles~\cite{estier2000shrimp}. However, passive suspension mechanisms cannot be optimized for all terrain conditions simultaneously. A design tailored to a particular obstacle geometry is likely to underperform in others~\cite{nayar2019design}. Even when considering a single obstacle type, such as a step, there exists a lower bound on the friction coefficient below which no passive suspension can succeed, despite the diversity of proposed designs~\cite{thueer2006comprehensive}. This limitation arises because passive systems rely exclusively on wheel–ground traction to generate the forces required for obstacle negotiation. On soft soil, the limitations of passive suspensions become particularly acute. Despite conservative path planning based on orbital imagery, NASA’s Opportunity rover experienced severe wheel sinkage at Endeavour Crater~\cite{arvidson2011opportunity}, while its twin Spirit became irretrievably embedded in soft sand, ultimately leading to mission termination~\cite{sanderson2010mars}. Reflecting this well-recognized vulnerability, the Curiosity rover enforces operational constraints that limit traverses to slopes below \qty{15}{\degree}~\cite{heverly2013traverse}.

One approach to extending mobility is to augment otherwise passive rovers with limited active reconfiguration. For example, actuators originally introduced for wheel deployment can be repurposed to shift the center of mass of six-wheeled rovers during obstacle traversal~\cite{patel2010exomars,skonieczny2010improving}. Similarly, four-wheeled platforms such as the Sample Return Rover and Scarab combine passive bogies with a small number of additional actuators to modulate chassis height and roll attitude~\cite{iagnemma2000mobile,wettergreen2010design}. These designs improve adaptability while avoiding the complexity of fully actuated limbs. However, their ability to redistribute load remains limited: they cannot control load transfer between front and rear wheels, and they rely on global body inclination, which is inherently coupled to vehicle stability. Moreover, they do not allow independent control of load distribution and wheel placement, which is required to actively sequence and assist obstacle traversal. Combining joints about the yaw and roll axes, the platform OpenWHEEL introduced the use of coordinated wheel repositioning and roll actuation to sequentially negotiate step-like obstacles~\cite{fauroux2006new}. However, the proposed control strategy remained open-loop and did not incorporate sensing feedback. Furthermore, the yaw joints, located above each wheel axle, are passive and driven by differential wheel speeds, and thus cannot actively assist in generating the forces required for obstacle negotiation.

At the opposite end of the spectrum, fully actuated wheel-on-limb systems provide a significantly richer locomotion repertoire by enabling each wheel to be positioned independently with respect to the chassis~\cite{leppanen1998workpartner,grand2004stability,wilcox2007athlete,reid2016actively,cordes2018design,reid2021actively}. Such platforms can achieve highly capable behaviors, including obstacle surmounting strategies that approach those of legged systems. However, they are constrained by the control complexity required to exploit their full kinematic potential. In addition, the large number of joints is seldom fully utilized, meaning that much of the associated actuation contributes primarily as mass and mechanical overhead.

For locomotion on soft soils, a wheel can generate greater thrust when operated in a pushing configuration than through traction alone~\cite{bekker1969introduction}. Consequently, several platforms have sought to exploit stationary or quasi-stationary wheels to assist propulsion through additional chassis actuation. This principle led to the development of ``rolling-peristaltic'' locomotion in the highly actuated six-wheeled Marsokhod chassis~\cite{andrade1998modeling}. Similarly, suspension-assisted locomotion has been demonstrated on a six-wheeled ExoMars test rover, where deployment actuators within its 16-actuator suspension system enable ``wheel-walking'' behaviors in sandy terrain~\cite{azkarate2015first}. In the case of four-wheeled platforms, fully actuated wheel-on-limb architectures such as WorkPartner and Hylos achieve ``rolking''~\cite{halme2000hybrid} and ``peristaltic symmetric''~\cite{amar2004performance} gaits, respectively. Using suspension kinematics derived from NASA’s RP15 lunar rover, the Mini Rover demonstrated a gait capable of effectively ``swimming'' up loosely consolidated slopes~\cite{shrivastava2020material}. However, this motion requires eight additional active joints beyond the wheel drive actuators, along with a four-bar linkage at each wheel. With only two actuated joints controlling the wheelbase length of each bogie, Scarab achieves a ``push–pull'' or ``inch-worming'' motion~\cite{wettergreen2010design,creager2015push}. The rover MARCEL, also equipped with two actuated joints but oriented orthogonally to each other, enables a ``crawling'' motion~\cite{bouton2022crawling} that combines wheel reconfiguration with active load redistribution. The latter approach was found experimentally to be more energy-efficient on steep, unconsolidated sandy slopes than the aforementioned gaits~\cite{bouton2023experimental}.

Model-free reinforcement learning (RL) has emerged as an effective framework for addressing complex locomotion problems. In recent years, it has achieved notable success in legged robotics~\cite{gurram2025reinforcement}, enabling the synthesis of controllers that are robust, versatile, and computationally efficient for real-time operation. For example, ANYmal has demonstrated the ability to combine multiple learned behaviors to traverse different types of obstacles, albeit relying on explicit obstacle classification and mode switching~\cite{hoeller2024anymal}. For wheel-on-limb platforms, the controller developed by Lee et al. leverages the full mobility of each limb to step over obstacles or climb stairs~\cite{lee2024learning}. Building on~\cite{miki2022learning}, their approach relies on a privileged teacher--student distillation framework coupled with a learned belief state, explicitly designed to cope with partial and noisy terrain perception. However, the terrain is assumed to be rigid throughout, with Coulomb friction; therefore, the agent is not exposed to the limitations of wheeled locomotion on soft soil. In wheeled systems with limited actuation, the rover MARCEL showed that a two-degree-of-freedom active chassis can be trained via reinforcement learning to negotiate step-like obstacles, although the study was restricted to a single scenario~\cite{bouton2023marcel}.

\subsection{CONTRIBUTIONS}

The contributions of this paper are fourfold.

First, it presents the design of the ERNEST planetary rover, integrating a simple yet versatile Active Gimbal Suspension, a clutch mechanism enabling transitions between passive and active suspension modes, and an associated wheel-synchronization control scheme for kinematically constrained steering and reconfiguration.

Second, it develops a reinforcement-learning framework using the high-fidelity DARTS simulation engine, which combines rigid contact and soft-soil terramechanics, enabling validation of a previously theorized gait through the observation of emergent optimal behavior.

Third, it introduces a policy-consolidation strategy that merges terrain-specialized experience into a single neural network, eliminating the need for multiple controllers and explicit switching.

Fourth, it demonstrates deployment of the unified controller on the physical rover across diverse terrains, including rock fields, a Bickler trap, a wheel-high step, sand ripples, and both dry and wet unconsolidated sandy slopes, showing that the learned policy exhibits interpretable behaviors such as sequential wheel climbing, active load redistribution, and crawling-like gaits.

\subsection{CONTENT}

The remainder of this paper is organized as follows. Section~\ref{sec:ernest} presents the ERNEST rover, its sensing and actuation architecture, the principles of the Active Gimbal Suspension, and the control approach used to coordinate the wheels. Section~\ref{sec:simulation} describes the DARTS simulation environment, the soft-soil terramechanics implementation, and the system identification procedures used to align the models with real-world behavior, facilitating transfer to the physical rover. Section~\ref{sec:rl_framework} details the reinforcement-learning formulation, the policy-consolidation procedure, and the definition of the different terrain classes used for training. Section~\ref{sec:results} reports the behavior of the resulting controller on the physical rover across a variety of obstacles and compares its performance with that of a passive bogie configuration.

A supplementary video accompanying this paper is available at \url{https://youtu.be/d684P5a3xMc}.

\section{ERNEST}
\label{sec:ernest}

ERNEST is a \qty{75}{\kilogram}, four-wheeled planetary rover concept equipped with an actively articulated chassis designed to improve mobility on challenging terrain, including obstacles that exceed the capabilities of conventional four- or six-wheeled passive suspension systems.
The acronym ERNEST stands for \emph{Exploration Rover for Navigating Extreme Sloped Terrain}. Its main dimensions are summarized in Table~\ref{tab:dimensions}.

\begin{figure}
	\centering
	\includegraphics[width=0.35\textwidth]{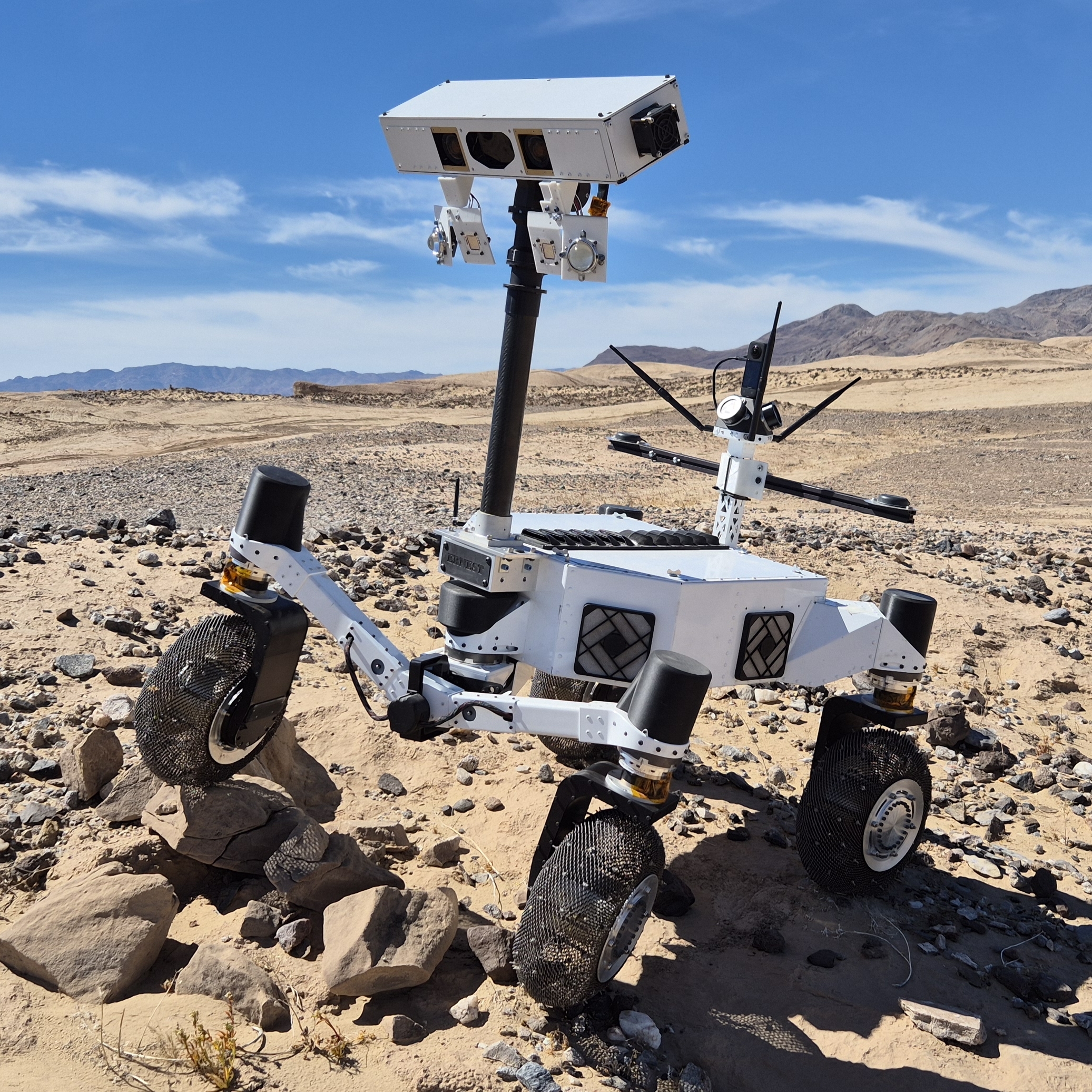}
	\caption{The ERNEST rover in the Yuha Desert, CA.}
	\label{fig:ernest_desert}
\end{figure}

\begin{table}
    \centering
    \begin{tabular}{lc}
        \toprule
        Wheelbase                       & \qty{86}{\cm} \\
        Wheel track                     & \qty{80}{\cm} \\
        Wheel diameter                  & \qty{34}{\cm} \\
        Wheel width                     & \qty{13}{\cm} \\
        Masthead height from the ground & \qty{130}{\cm} \\
        Clearance under the bogie       & \qty{30}{\cm} \\
        Clearance under the chassis     & \qty{42}{\cm} \\
        \hline
    \end{tabular}
    \vspace{1.5ex}
    \caption{Main dimensions of ERNEST.}
    \label{tab:dimensions}
\end{table}

\subsection{SYSTEM OVERVIEW}

\subsubsection{Mobility Hardware}

ERNEST features ten actuators dedicated to mobility: four wheel drives, four steering actuators located above each wheel assembly, and two actuators for the Active Gimbal Suspension (AGS). All actuators are identical in design, each consisting of a brushless DC motor equipped with a power-on-to-disengage electromagnetic brake, Harmonic Drive reduction, and two absolute magnetic encoders associated with the input and output stages. The only distinction lies in the gear ratios: the drive and steering actuators use a 50{:}1 reduction, whereas the AGS actuators use a 160{:}1 reduction.

Although ERNEST is inherently capable of omnidirectional motion through its four steering actuators, these are not utilized in this work. Instead, the front and rear wheel pairs remain aligned with their respective axle at all times, and steering is achieved exclusively through the AGS, effectively reducing the number of active mobility actuators to six.

ERNEST's wheels consist of compliant, airless tires composed of hundreds of woven coiled steel springs that form a toroidal mesh. These tires are designed to deform to rugged terrain, providing the rover with improved traction and resistance to punctures compared to rigid wheels~\cite{asnani2009development,lu2026design}. Microspikes are riveted to the mesh sparsely along the wheel circumference to enhance the wheel's ability to maintain grip and traction when navigating rocky surfaces or climbing obstacles.

\subsubsection{Sensors}

\begin{figure}
	\centering
	\includegraphics[width=0.45\textwidth]{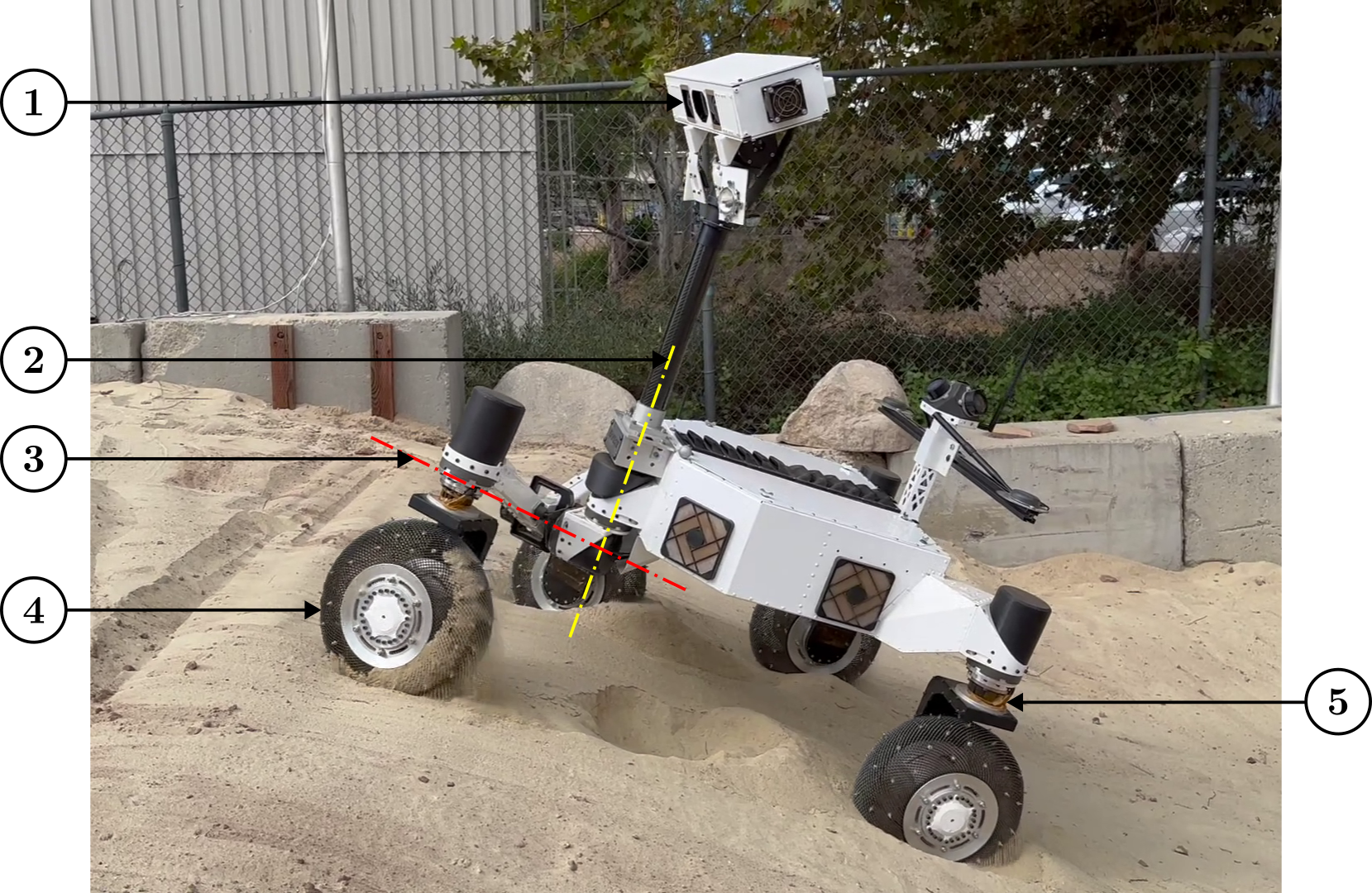}
	\caption{Components of the ERNEST rover used in this work: (1) stereo cameras; (2) yaw-joint axis; (3) roll-joint axis; (4) meshed wheels; and (5) force–torque sensors.}
	\label{fig:ernest_slope}
\end{figure}

As shown on Fig.~\ref{fig:ernest_slope}, force--torque sensors are mounted above each wheel assembly, just under the steering actuators, to provide feedback on wheel--ground interaction forces.

Two cameras housed in the masthead provide stereo vision, terrain mapping, and visual–inertial odometry. An inertial measurement unit (IMU) is positioned between them to support this estimation. Illuminators for night driving are suspended from the masthead below the cameras. Although the mast is equipped with two brushed DC motors enabling pan–tilt motion of the masthead, it is kept fixed in this work, oriented straight ahead and \qty{30}{\degree} downward relative to the chassis horizontal.

\begin{figure}
	\centering
	\includegraphics[width=0.48\textwidth]{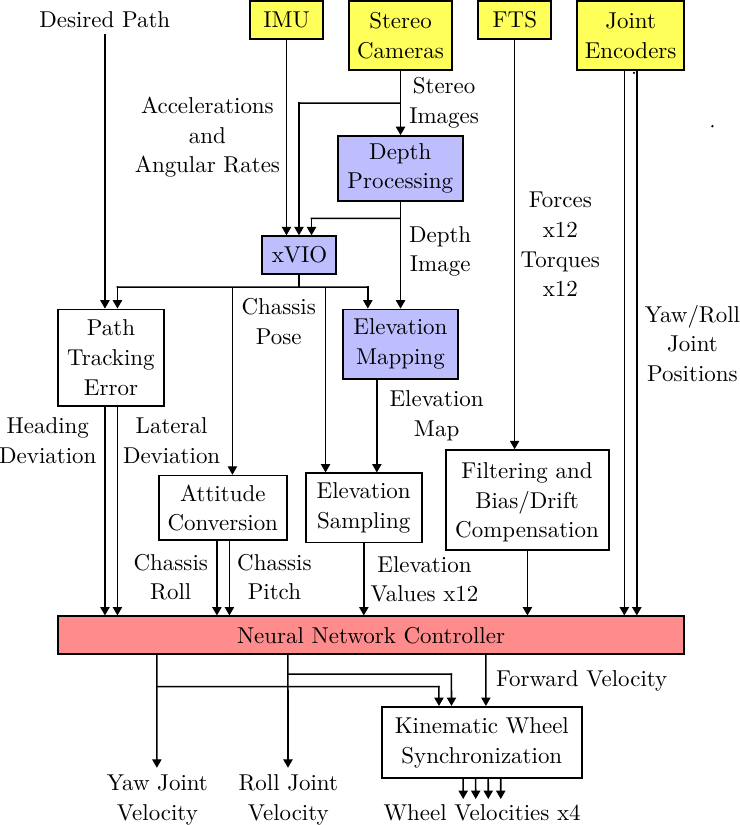}
	\caption{Diagram of the perception and control architecture. The sensors are highlighted in yellow. The blue blocks are adapted from prior work.}
	\label{fig:system_diagram}
\end{figure}

The sensors employed and the processing pipeline of their associated data products are illustrated in Fig.~\ref{fig:system_diagram}.

The rover is also equipped with additional sensors that are not used in this work: a LiDAR unit housed within the masthead, sun sensors on top of the rear mast, two GPS antennas located on either side of that mast, and a second IMU within the chassis.

\subsubsection{Avionics}

The embedded computer housed within the chassis is an Intel NUC, a compact high-performance x86 platform based on an Intel Core i7 processor. The actuators are driven by Elmo Platinum Solo Twitter servo drives, also integrated within the chassis to benefit from its forced-air cooling in a positive-pressure enclosure. The onboard computer communicates with the drives over an EtherCAT bus, which additionally interfaces with a set of Beckhoff I/O terminals. These terminals support interfacing functions including energizing the clutch shotbolt solenoids, controlling the masthead illuminators, reading the temperature sensors installed on the motor windings, and communicating with the independent encoder measuring the relative angle between the bogie and its frame (see Fig.~\ref{fig:clutch}). The force–torque sensors are also connected to the main daisy-chained EtherCAT segment through a hub that distributes the bus to each of the four wheel assemblies in a star topology. Remote access to the onboard computer is provided via a router integrated within the chassis, with Wi-Fi antennas mounted on a dedicated mast at the rear of the rover.

\subsubsection{Power}

The rover is powered by custom \qty{82}{\volt}, \qty{17.2}{\ampere\hour} lithium-ion battery packs installed within the chassis. It can operate continuously for up to two hours with a single pack, or up to four hours with two packs in parallel. The mobility actuators are supplied directly from the \qty{82}{\volt} bus, while a power distribution board steps this bus down to regulated \qty{24}{\volt}, \qty{19}{\volt}, and \qty{12}{\volt} rails for the different subsystems operating at those voltage levels.

\subsection{ACTIVE GIMBAL SUSPENSION}

ERNEST’s suspension consists of a two-degree-of-freedom actively controlled joint assembly connecting the front wheel bogie to the main chassis. As shown in Fig.~\ref{fig:ernest_slope}, the two revolute joints are collocated, with intersecting axes, forming a gimbal-like mechanism referred to as the \emph{Active Gimbal Suspension}. One revolute joint provides yaw motion of the front bogie about a vertical axis relative to the chassis, located near the mast axis, while the second revolute joint enables roll motion about the bogie’s own local longitudinal axis.

This combination of joints was chosen for its effective trade-off between mechanical simplicity and functional capability, as demonstrated in previous studies~\cite{bouton2022crawling,bouton2023marcel,bouton2023experimental}. However, their collocation in ERNEST's variation of the design is primarily motivated by thermal management considerations, enabling most of the avionics to be housed within a single chassis enclosure.

The yaw joint alone enables slip-free steering under ideal kinematic assumptions when the wheel-synchronization control presented in the next section is applied.
This joint also allows the rover to reconfigure its wheel positions when negotiating obstacles, thereby effectively serving two functions with a single actuator.

\begin{figure}
	\centering
	\newcommand{\widthratio}{0.45}

	\renewcommand{\thesubfigure}{a}%
	\subfloat[]{%
        \includegraphics[width=\widthratio\textwidth]{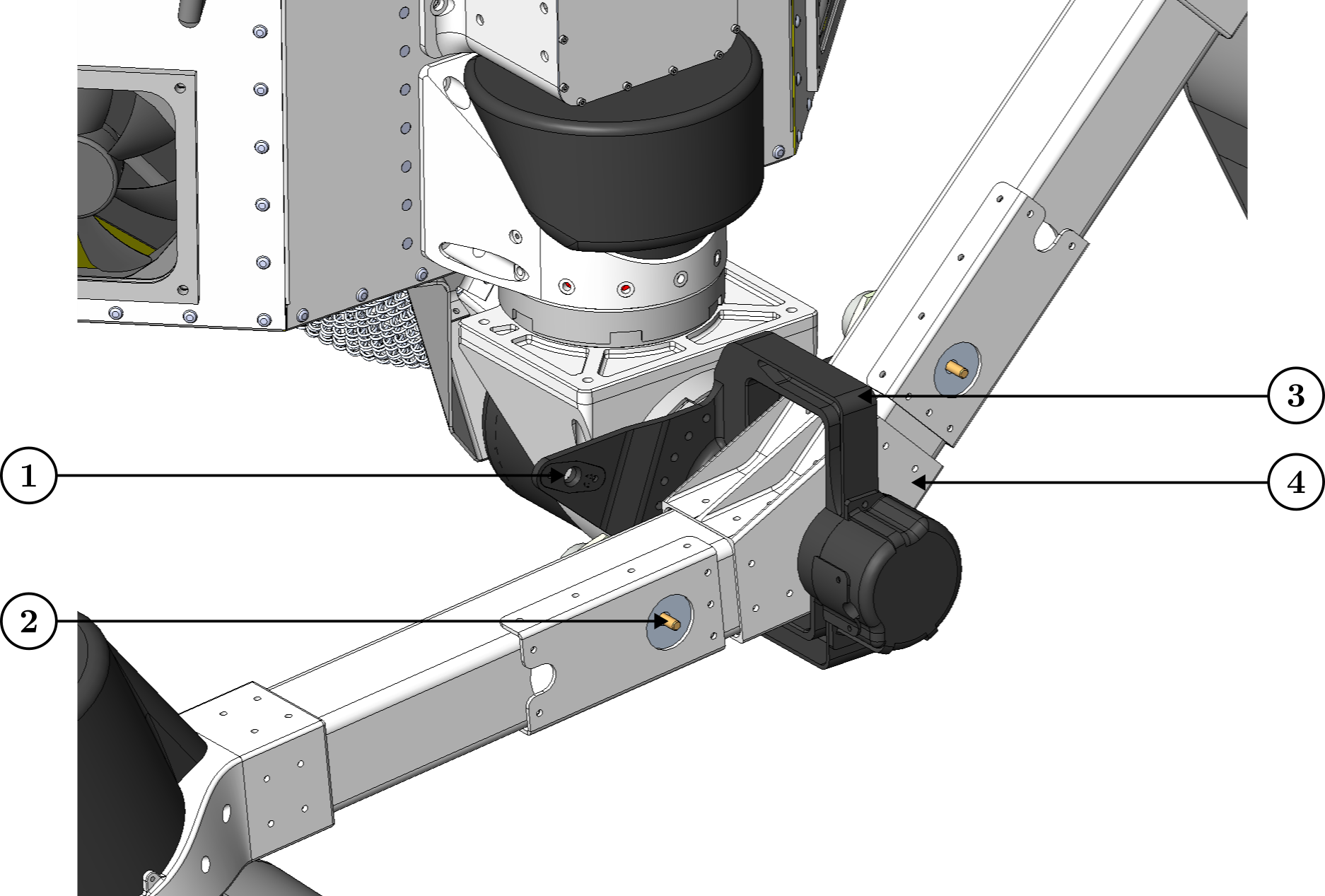}%
        \label{fig:clutch_disengaged}%
    }
	\hfil
	\renewcommand{\thesubfigure}{b}%
	\subfloat[]{%
        \includegraphics[width=\widthratio\textwidth]{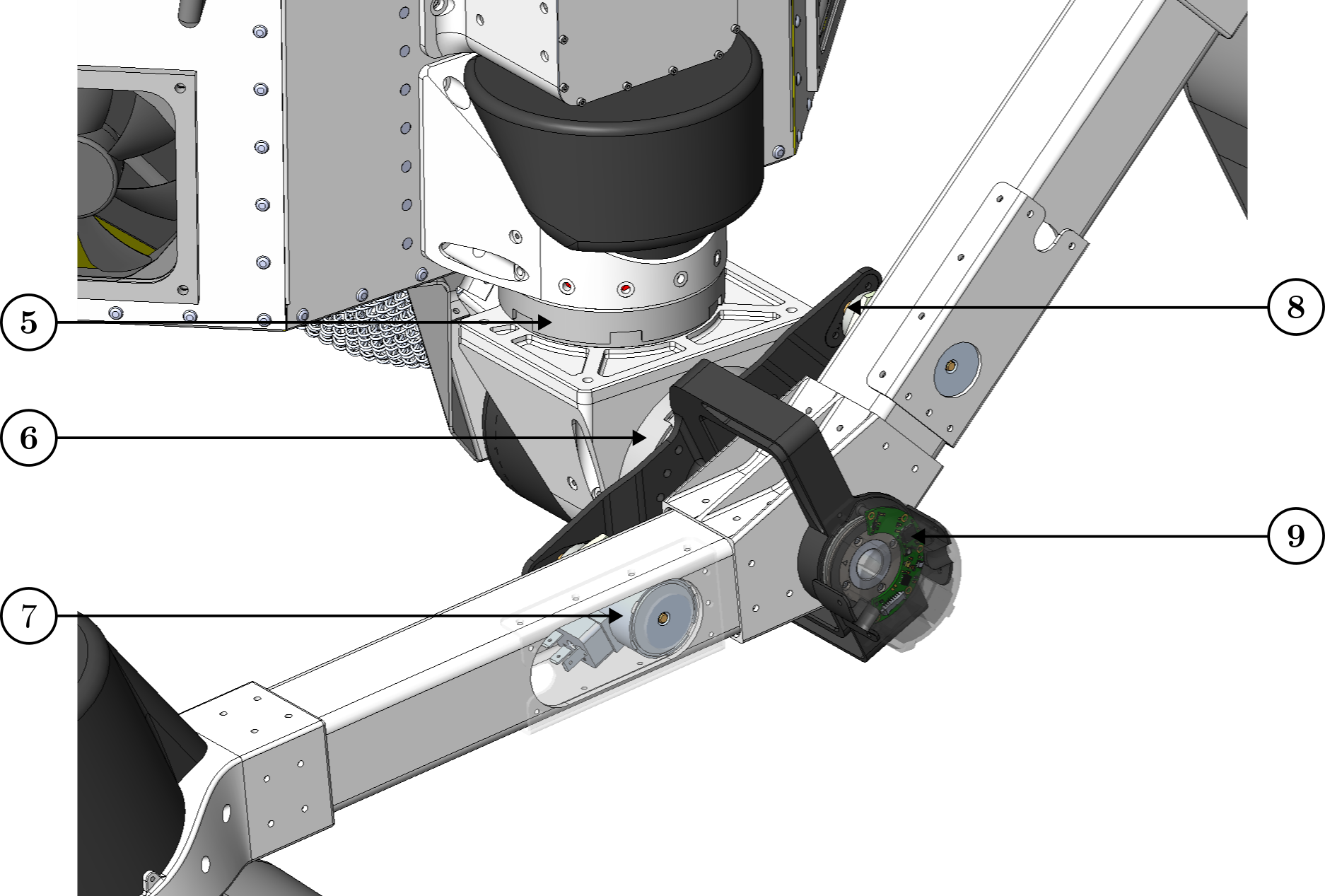}%
        \label{fig:clutch_engaged}%
    }

	\caption{The clutch mechanism of the Active Gimbal Suspension. \protect\subref{fig:clutch_disengaged} Clutch disengaged: the bogie is free to rotate relative to the bogie frame. \protect\subref{fig:clutch_engaged} Clutch engaged: the rotation of the bogie is directly controlled by the roll actuator. (1) Pin hole. (2) Retracted pin of the shotbolt lock unit. (3) Bogie frame. (4) Bogie connected to the bogie frame through a pivot joint supported by ball bearings. (5) Output of the yaw actuator. (6) Output of the roll actuator. (7) Solenoid of the shotbolt lock unit inside the arm of the bogie. (8) Extended pin locked into the pin hole of the bogie frame. (9) Absolute magnetic encoder reporting the relative angle between the bogie and the bogie frame.}
	\label{fig:clutch}
\end{figure}

The roll joint can either rotate freely or be actively controlled. Coupling between the roll actuator and the bogie is achieved through a clutch mechanism composed of two shotbolt lock units embedded in the bogie arms, as show in Fig.~\ref{fig:clutch}. When the pins of these two solenoids are retracted, the bogie rotates freely relative the bogie frame through a set of ball bearings. This configuration minimizes energy consumption when the terrain is benign enough to be traversed with a passive suspension that naturally conforms to the ground geometry under gravity.
The bogie frame is itself rigidly connected to the output of the roll actuator and can be actively rotated independently of the bogie motion when the clutch is disengaged. Thus, the bogie frame can be rotated by the roll actuator at any time to align its pin holes with the solenoid pins in the bogie arms, as depicted by the transition between Fig.~\ref{fig:clutch_disengaged} and Fig.~\ref{fig:clutch_engaged}. Alignment is achieved by driving the relative angle between the bogie frame and the bogie, as measured by a dedicated absolute encoder, to zero using a velocity-feedback controller with a bounded maximum angular velocity. The controller enables reliable alignment even when the bogie undergoes motion induced by uneven terrain. Once aligned, the solenoids are energized to engage the pins with the bogie-frame inserts, enabling direct transmission of roll-actuator torque to the bogie.

When actively applying torque, the roll joint enables the rover to selectively shift the load distribution between the two diagonally opposite pairs of wheels. This can be used to facilitate the traversal of obstacles for which the available adhesion is insufficient, as demonstrated analytically in~\cite{bouton2023marcel}. This also allows the rover to lift one of its wheels off the ground. The wheel that is lifted for a given direction of rotation of the roll joint depends on the current rover configuration, as determined by the yaw joint position.
Thus, with only two joints, the Active Gimbal Suspension allows the rover to lift any of its wheels, using the yaw joint to move its bogie in a configuration that places its center of mass above the three other wheels, while the roll joint provides the lifting torque.

\subsection{WHEEL SYNCHRONIZATION}
\label{sec:wheel_synchronization}

The velocity of each wheel is computed according to the desired forward velocity of the chassis $V_x$, the velocities of the yaw and roll joints of the active gimbal suspension, denoted respectively $\dot{\psi}$ and $\dot{\phi}$, and the current configuration of the rover. Let $v^i_x$, $v^i_y$, and $v^i_z$ be the components of the translational velocity of the center of the wheel $i$ expressed in the local frame of the wheel where the $y$-axis is aligned with its axis of rotation. Given that the steering joints above the wheels are not used here and locked such that each pair of front or rear wheel axles are aligned, the Jacobian matrix for the motion of one wheel $i$ can be expressed as:

\begin{equation}
  \mathbf{J_i\dot{q}} =
  \left[\renewcommand{\arraystretch}{1.1}
  \begin{array}{c}
    v^i_x \\
    v^i_y \\
    v^i_z
  \end{array}
  \right],
\end{equation}
where $\mathbf{J_i}$ is made of the following column vectors:

\begin{equation}
\medmath{
  \mathbf{J_i} =
  \begin{bmatrix}
    \mathbf{C}^i_{V_x} & \mathbf{C}^i_{V_y} & \mathbf{C}^i_{V_z} & \mathbf{C}^i_{\omega_x} & \mathbf{C}^i_{\omega_y} & \mathbf{C}^i_{\omega_z} & \mathbf{C}^i_{\dot{\psi}} & \mathbf{C}^i_{\dot{\phi}}
  \end{bmatrix}
}.
\end{equation}
These column vectors each corresponds to a generalized velocity coordinate of $\dot{q}$, which consists of:

\begin{equation}
  \mathbf{\dot{q}} =
  \begin{bmatrix}
    V_x & V_y & V_z & \omega_x & \omega_y & \omega_z & \dot{\psi} & \dot{\phi}
  \end{bmatrix}^\mathsf{T},
\end{equation}
with $V_x$, $V_y$, and $V_z$ the components of the translational velocity of the chassis in the local frame of the chassis, in which the $x$-axis coincides with the longitudinal axis of the main body of the rover, orthogonal with the rear wheel axles and pointing forward. $\omega_x$, $\omega_y$, and $\omega_z$ are the components of the rotational velocity of the chassis in that same frame, while $\dot{\psi}$ and $\dot{\phi}$ are the velocities of respectively the yaw and roll joints of the active gimbal suspension.

Using the column vectors of the Jacobian matrices for each wheel numbered from 1 to 4 and denoted as superscripts in the equations below, we can compose the following system of equations:

\begin{equation}
  \mathbf{Ax} = \mathbf{y},
\end{equation}
where $\mathbf{x}$ is the vector of unknown variables that we are solving for:

\begin{equation}
\medmath{
  \mathbf{x} =
  \left[\begin{array}{ccc|cc|cc|cc|cc}
    V_y & V_z & \omega_z & v^1_x & v^1_z & v^2_x & v^2_z & v^3_x & v^3_z & v^4_x & v^4_z
  \end{array}\right]^\mathsf{T}
}.
\end{equation}
The matrix $\mathbf{A}$ can then be written:
\begin{equation}
\medmath{
  \mathbf{A} =
  \left[\renewcommand{\arraystretch}{1.2}\begin{array}{ccc|c|c|c|c}
    \mathbf{C}^1_{V_y} & \mathbf{C}^1_{V_z} & \mathbf{C}^1_{\omega_z} & -\mathbf{I_{3x2}} & \mathbf{0_{3x2}} & \mathbf{0_{3x2}} & \mathbf{0_{3x2}} \\
    \mathbf{C}^2_{V_y} & \mathbf{C}^2_{V_z} & \mathbf{C}^2_{\omega_z} & \mathbf{0_{3x2}} & -\mathbf{I_{3x2}} & \mathbf{0_{3x2}} & \mathbf{0_{3x2}} \\
    \mathbf{C}^3_{V_y} & \mathbf{C}^3_{V_z} & \mathbf{C}^3_{\omega_z} & \mathbf{0_{3x2}} & \mathbf{0_{3x2}} & -\mathbf{I_{3x2}} & \mathbf{0_{3x2}} \\
    \mathbf{C}^4_{V_y} & \mathbf{C}^4_{V_z} & \mathbf{C}^4_{\omega_z} & \mathbf{0_{3x2}} & \mathbf{0_{3x2}} & \mathbf{0_{3x2}} & -\mathbf{I_{3x2}} \\
  \end{array}\right]
},
\end{equation}
with:
\begin{align}
  \mathbf{I_{3x2}} =
  \begin{bmatrix}
    1 & 0 \\
    0 & 0 \\
    0 & 1 \\
  \end{bmatrix},
  &&
  \mathbf{0_{3x2}} =
  \begin{bmatrix}
    0 & 0 \\
    0 & 0 \\
    0 & 0 \\
  \end{bmatrix}.
\end{align}
The variables that are known are moved to the right-hand side of the equation such that:
\begin{equation}
\medmath{
  \mathbf{y} =
  -\begin{bmatrix}
    \mathbf{C}^1_{V_x} \\ \mathbf{C}^2_{V_x} \\ \mathbf{C}^3_{V_x} \\ \mathbf{C}^4_{V_x}
  \end{bmatrix}V_x
  -\begin{bmatrix}
    \mathbf{C}^1_{\omega_x} \\ \mathbf{C}^2_{\omega_x} \\ \mathbf{C}^3_{\omega_x} \\ \mathbf{C}^4_{\omega_x}
  \end{bmatrix}\omega_x
  -\begin{bmatrix}
    \mathbf{C}^1_{\omega_y} \\ \mathbf{C}^2_{\omega_y} \\ \mathbf{C}^3_{\omega_y} \\ \mathbf{C}^4_{\omega_y}
  \end{bmatrix}\omega_y
  -\begin{bmatrix}
    \mathbf{C}^1_{\dot{\psi}} \\ \mathbf{C}^2_{\dot{\psi}} \\ \mathbf{C}^3_{\dot{\psi}} \\ \mathbf{C}^4_{\dot{\psi}}
  \end{bmatrix}\dot{\psi}
  -\begin{bmatrix}
    \mathbf{C}^1_{\dot{\phi}} \\ \mathbf{C}^2_{\dot{\phi}} \\ \mathbf{C}^3_{\dot{\phi}} \\ \mathbf{C}^4_{\dot{\phi}}
  \end{bmatrix}\dot{\phi}.
}
\label{eq:y}
\end{equation}
On the right-hand side of the equation, the lateral wheel velocities $v^1_y$, $v^2_y$, $v^3_y$, and $v^4_y$ would normally appear. However, to impose a no-side-slip constraint at the wheels, thereby allowing only for tangential motion, these terms are all zero and are therefore omitted from $\mathbf{y}$.

To solve this system while minimizing the quadratic sum of the velocities in $\mathbf{x}$, we use the right pseudoinverse of the matrix $\mathbf{A}$:

\begin{equation}
  \mathbf{x} = \mathbf{A^+ y}.
\end{equation}
This yields the tangential velocity components of each wheel. Given the wheel radius $r$, we can finally deduce the angular velocity of each wheel $i$ as:
\begin{equation}
  \omega^i_{wheel} = \frac{1}{r}\sign(v^i_x)\sqrt{{v^i_x}^2 + {v^i_z}^2}.
\end{equation}






\subsection{PERCEPTION PIPELINE}

The rover employs an onboard stereo perception pipeline to estimate its state and reconstruct local terrain geometry. As shown in Fig.~\ref{fig:system_diagram}, both the estimated rover pose and the terrain elevation map are subsequently used by the control scheme.

The rover’s 6-DoF state and associated uncertainty are estimated at \qty{4}{\hertz} using the stereo–inertial odometry algorithm xVIO described in~\cite{delaune2020xvio}. xVIO employs a tightly coupled extended Kalman filter (EKF) formulation that fuses high-rate pre-integrated inertial measurements with sparse visual feature tracks from the stereo cameras. This approach provides robust, low-drift state and covariance estimation even in challenging terrain. To perceive the local environment, dense 3D point clouds are generated using the OpenCV stereo block-matching (StereoBM) implementation~\cite{Konolige1998}.

Stereo point clouds are fused at \qty{0.5}{\hertz} using the corresponding xVIO pose estimates to construct a robot-centric \qtyproduct{10x10}{\meter}, 2.5D elevation map at \qty{1}{\cm} resolution. This map is generated using the robot-centric Elevation Mapping software package~\cite{fankhauser2018probabilistic}. An example of the resulting elevation map is shown in Fig.~\ref{fig:elevation_sampling}. Rather than naively overwriting previous measurements, the framework explicitly models depth sensor noise and 6-DoF pose uncertainty, and employs a probabilistic Kalman filter–based update for each grid cell to continuously refine terrain height estimates and their associated variances.

\section{SIMULATION}
\label{sec:simulation}

\subsection{DARTS}

The Dynamics Algorithms for Real-Time Simulation (DARTS) framework is used to develop a simulated training environment. DARTS is a general multibody dynamics simulation framework developed in-house at JPL~\cite{bonilla2025dshell}. DARTS simulations have been developed for numerous mission domains at JPL, including spacecraft Guidance, Navigation, and Control (GNC), Entry-Descent and Landing (EDL), rotorcraft GNC, ground vehicle mobility, and robotic autonomy. DARTS is used in this work because it offers fast, flight validated, $\mathcal{O}(N)$ recursive dynamics formulations~\cite{jain2011book}, proven vehicle mobility models that could be readily extended to the ERNEST platform, and a flexible runtime terrain generation toolkit amenable to domain randomization~\cite{bonilla2026multifidelity}. 

The ERNEST multibody kinematic configuration, mass model (discussed later), and visual/collision mesh geometries are implemented in DARTS. Simulated DC motor and PID controller models are added to the wheel, steer, roll, and yaw joints. Force--Torque Sensor (FTS) models report the inter-body wrench between each steer body and their respective mounting point on the chassis, extracted directly from the DARTS multibody dynamics solver. Joint stops are simulated by applying a stiff spring-damper force.

\begin{figure}
	\centering
	\includegraphics[width=0.45\textwidth]{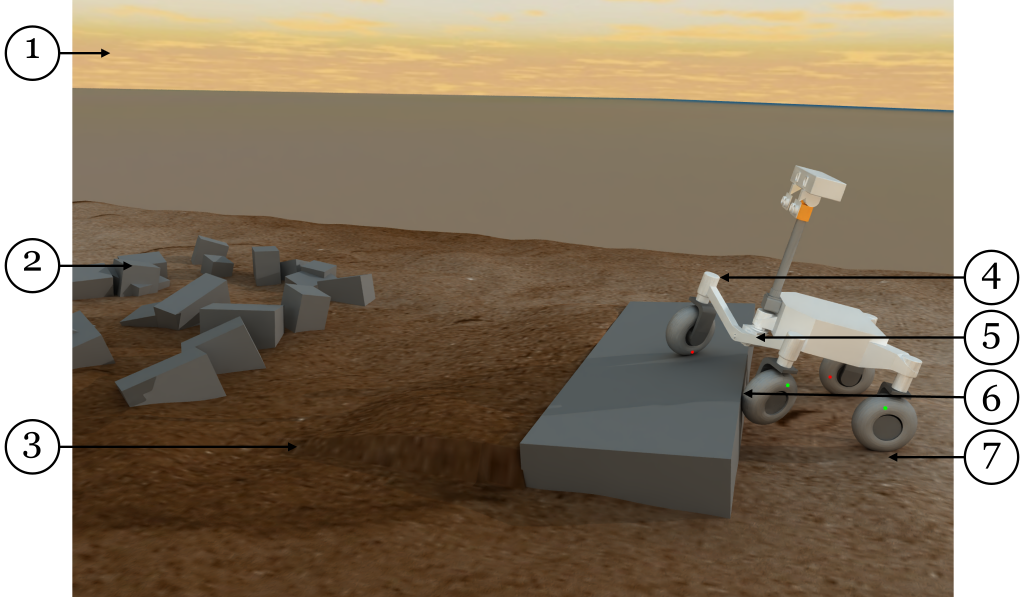}
	\caption{DARTS training environment. (1) GPU-accelerated path-traced rendering backend. This is used only for visualization in the present work but can readily simulate perception cameras and lidars if needed for follow-on work \cite{aiazzi2022IRIS}. (2) Placeable hard collision obstacles. (3) User-generated terrain with dispersed noise. (4) Controllable wheel/steer motors with encoder and Force--Torque Sensor (FTS) models. (5) Controllable ERNEST Active Gimbal Suspension (roll and yaw) with joint stops. (6) Stiff spring-damper contacts with isotropic friction between wheels and collision obstacles. (7) Soft soil terramechanics contacts between wheels and terrain. The terrain does not deform during these contacts.}
	\label{fig:darts_overview}
\end{figure}

The simulated training environment includes a base terrain represented by a Digital Elevation Model (DEM), generated by sampling a heightmap function: $h = f(x, y)$. Additional obstacles can be placed on the terrain. Fig.~\ref{fig:darts_overview} shows a summary of what is simulated during a typical training run.

The present work is among the first uses of DARTS for Reinforcement Learning (RL) training. Unlike some RL-focused simulation platforms, such as Isaac Lab developed by NVIDIA \cite{mittal2023nvidia}, DARTS does not support parallelizable dynamics that can be leveraged by GPU acceleration. However, multiple simulation processes can be run simultaneously on the CPU. High performance computing (HPC) resources are used to run many simulation processes concurrently. 

\subsection{MASS MODEL}
\label{sec:mass_model}

To ensure that the physical behavior of the rover and the force–torque sensor feedback remain consistent between simulation and real-world deployment, an accurate model of the rover’s three-dimensional mass distribution is required.

The rover is modeled as two rigid bodies connected by the Active Gimbal Suspension, which is represented as an ideal two-degree-of-freedom joint. The first body consists of the bogie and the two front wheels, while the second includes the remainder of the chassis, from the mast head to the rear wheels. Owing to the near symmetry of the bogie assembly with respect to the median and transverse planes, the center of mass (CoM) of this body is assumed to lie at the intersection of these two planes. The parameters to be identified are therefore the mass $m_b$ and CoM height $z_b$ of the bogie assembly, and the mass $m_c$ together with the three-dimensional CoM coordinates $(x_c, y_c, z_c)$ of the main chassis.

Under the assumption of zero torque at the roll joint, the system is statically determined, allowing the model to predict the vertical forces at each of the four wheel--ground contact points.

\begin{figure}
	\centering
	\newcommand{\widthratio}{0.22}

	\subfloat{\includegraphics[width=\widthratio\textwidth]{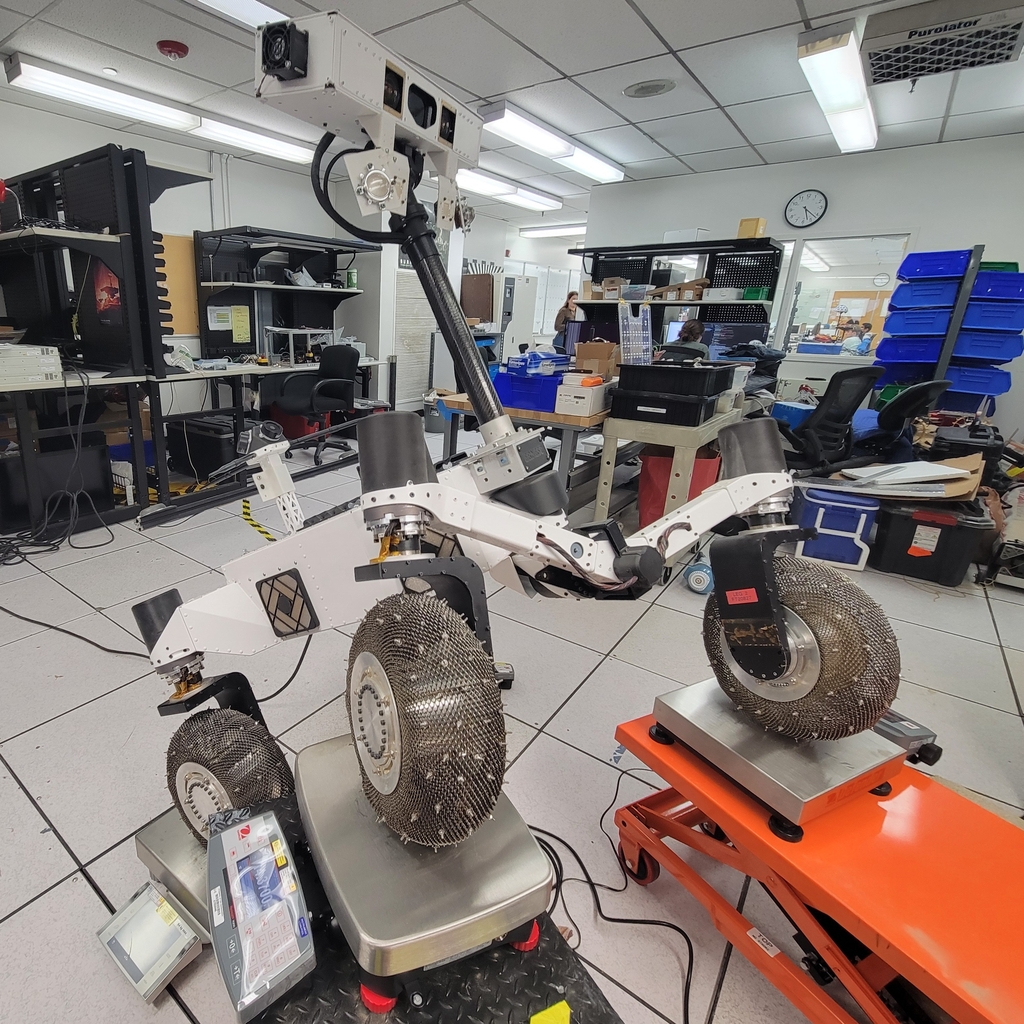}}
	\hfil
	\subfloat{\includegraphics[width=\widthratio\textwidth]{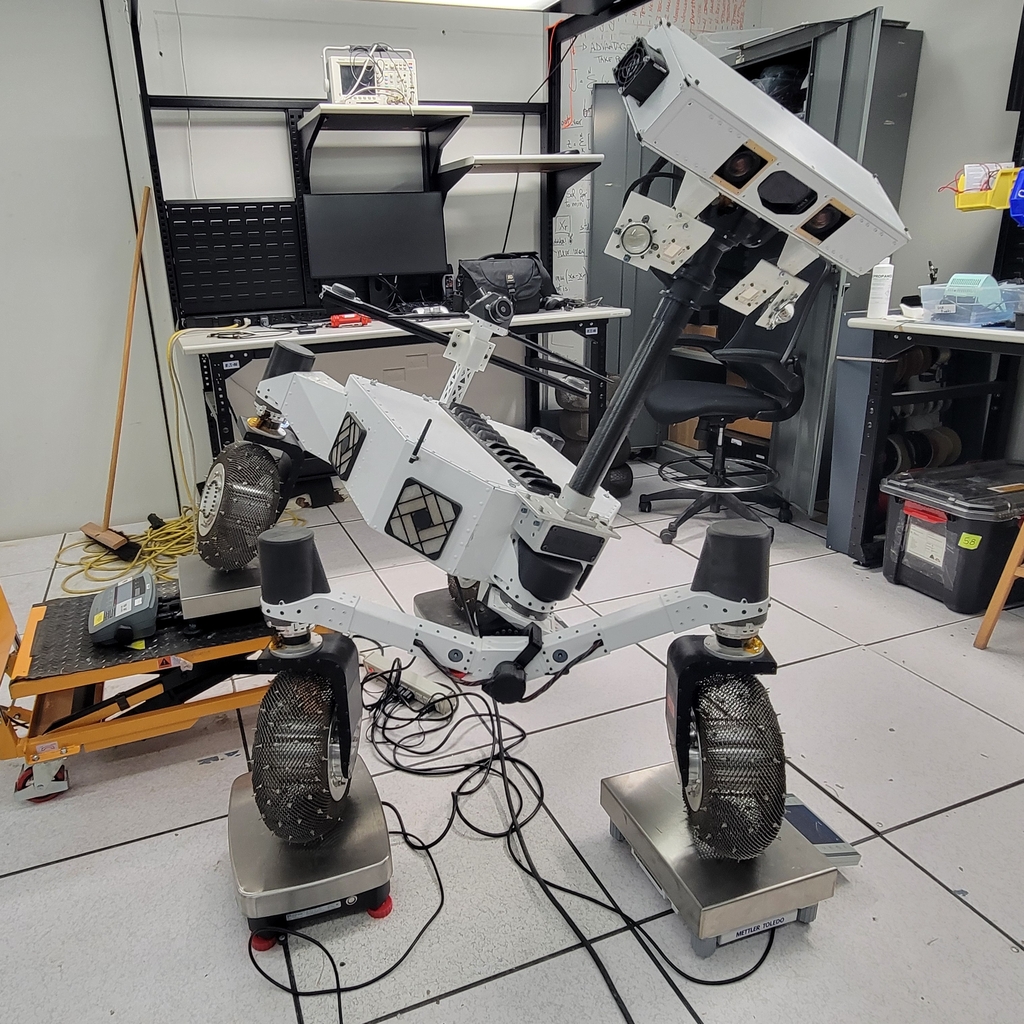}}

	\subfloat{\includegraphics[width=\widthratio\textwidth]{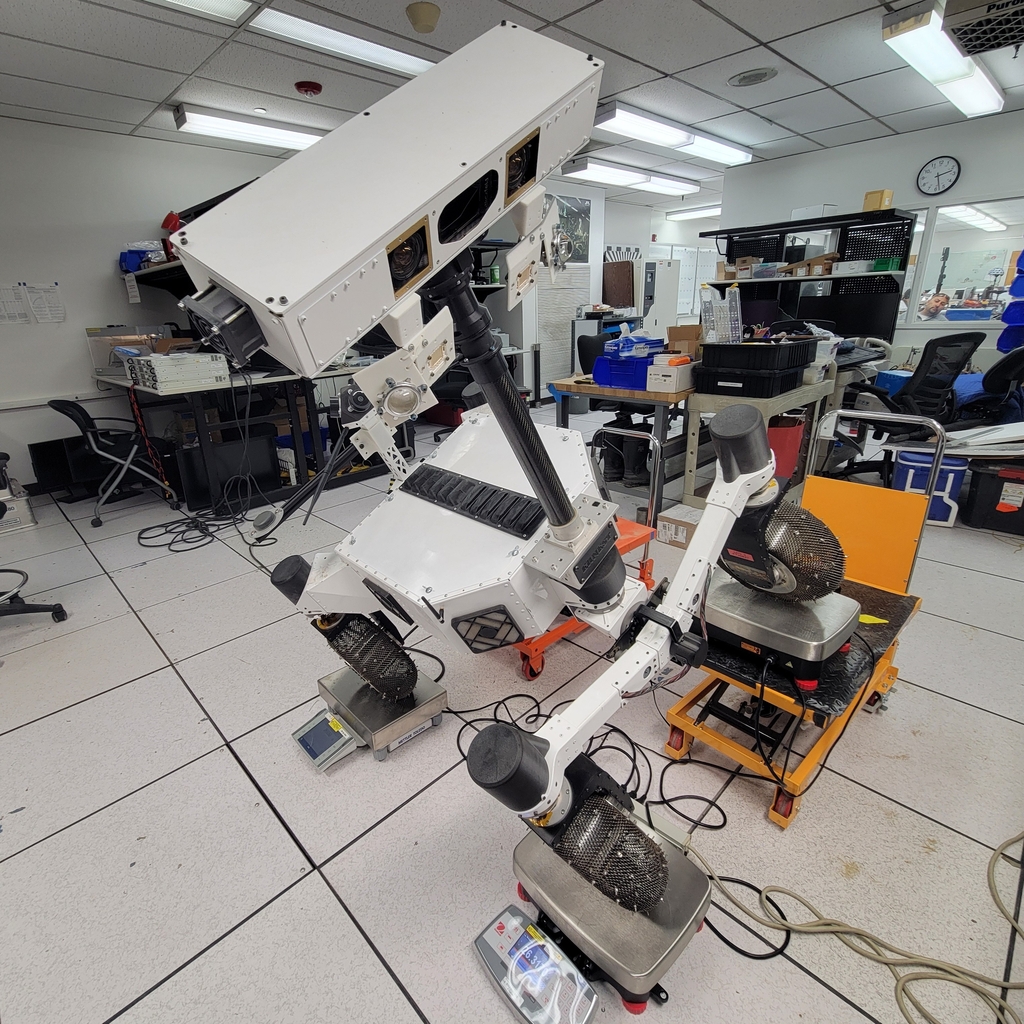}}
	\hfil
	\subfloat{\includegraphics[width=\widthratio\textwidth]{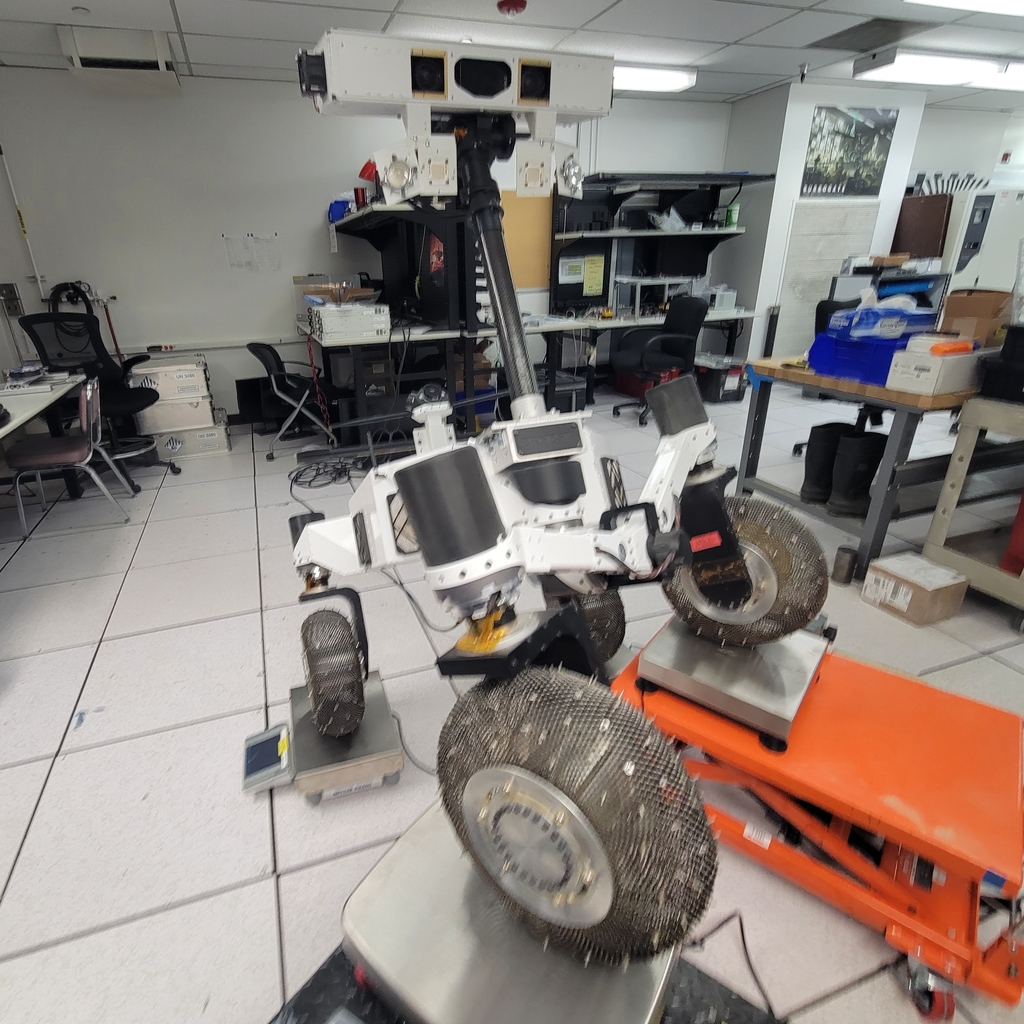}}

	\caption{Examples of rover configurations during the acquisition of load measurements for mass-distribution modeling.}
	\label{fig:mass_measurements}
\end{figure}

To identify the model parameters, the normal load under each wheel is measured experimentally using the physical rover over a wide range of configurations, spanning variations in chassis roll and pitch angles as well as in the positions of both joints of the Active Gimbal Suspension, as illustrated in Fig.~\ref{fig:mass_measurements}. During these measurements, the clutch is disengaged, allowing the bogie to rotate freely about the roll joint axis.

The model parameters are then estimated by nonlinear regression using the Levenberg–Marquardt algorithm to solve the following least-squares problem:
\begin{equation}
    \argmin_{
    \substack{
        m_c, m_b \\
        (x_c, y_c, z_c) \\
        z_b
    }
    }
    \sum_{\text{poses}} \sum_{i=1}^{4} \left( f_{z,i} - \hat{f}_{z,i} \right)^2,
\end{equation}
where $f_{z,i}$ denotes the measured vertical force at wheel $i$, and $\hat{f}_{z,i}$ is the corresponding force predicted by the mass distribution model. The resulting fit achieves a coefficient of determination of $R^2 = 0.96$.

\subsection{SOFT TERRAMECHANICS MODEL}


Contact between the rover's wheels and the terrain is treated using the semi-empirical Bekker-Wong soft soil terramechanics model. The implementation in DARTS is adapted from the approach in~\cite{ishigami2007wheel}, wherein derivations of the model in this section are provided. 

Implementing the Bekker-Wong model within a multibody vehicle simulation presents several challenges, primarily because the Bekker-Wong equations are often presented in a manner that is best-suited for simple, single-wheel performance analysis. The approach in~\cite{ishigami2007wheel} addresses some of these challenges, for example, adding a treatment of lateral shear stress for cornering. DARTS builds on this approach with additional extensions for multibody simulation. For example in a full-vehicle simulation, additional care is required to correctly determine the sign of various terms when the wheel is allowed move both forwards and backwards or when its motion is influenced by factors beyond wheel-soil interaction alone. Such scenarios occur when a rover mechanism (such as the active gimbal system on ERNEST) moves the wheel assembly, or when external forces like gravity cause the vehicle to slide down a slope.

Furthermore, the simulation must robustly handle scenarios where wheels can be at or near rest. Several terms in the formulation become ill-defined or discontinuous as wheel velocity approaches zero. To mitigate this, DARTS introduces regularization terms to smooth these transitions. This is critical when using adaptive-step numerical integrators as they are susceptible to instability or failure when encountering stiff or rapidly changing dynamics. DARTS incorporates several modifications to the equations in ~\cite{ishigami2007wheel} to address the above-mentioned complexities, which will be pointed out as they appear. 

\begin{figure}
	\centering
	\includegraphics[width=0.4\textwidth]{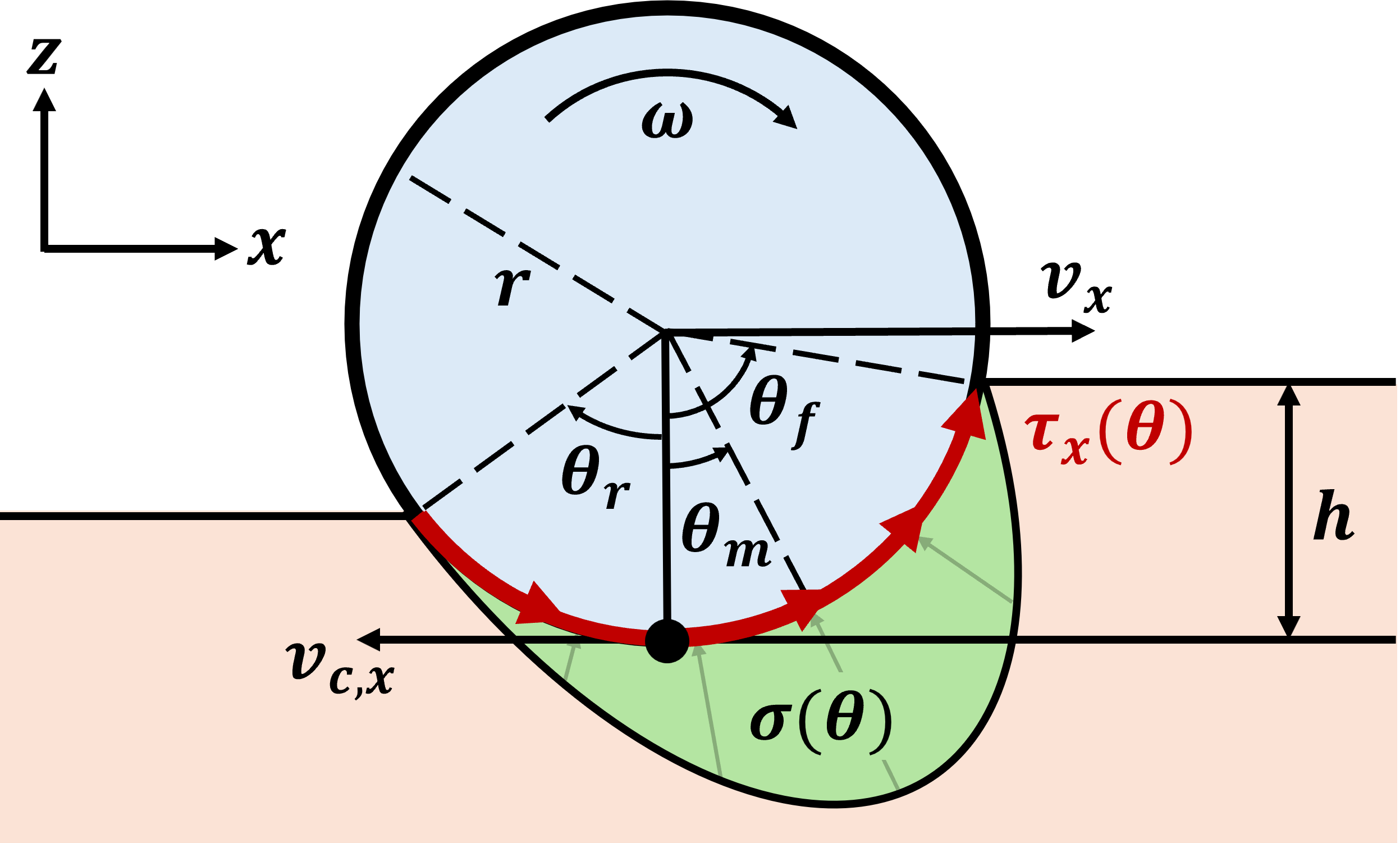}
	\caption{Geometric definitions for a wheel interacting with soft soil. The coordinate system is defined such that it yaws about $z$ with the wheel but does not rotate about $x$ or $y$ to match the wheel's tilt or spin. The $+x$ direction is oriented to be the forward direction of the vehicle. That is, $v_{x} < 0$ if the wheel moves backward to the left in the figure. Similarly the terrain-relative contact velocty $v_{c,x}$ is negative if moving to the left.  A positive wheel angular velocity $\omega$ is taken to be clockwise as shown in the figure. Finally, $\theta$ is taken to be zero at the bottom-most point of the wheel (large black circle in the figure) and is positive counter-clockwise. }
	\label{fig:bekker_diagram}
\end{figure}

Geometric definitions and sign conventions for a rigid cylindrical wheel of radius $r$ and width $b$ in contact with a soft soil are shown in Fig~\ref{fig:bekker_diagram}. A reference contact point is selected by taking the point on the wheel penetrating farthest into the terrain, shown as a large black circle in Fig~\ref{fig:bekker_diagram}. In DARTS the penetration depth $h$ and velocity of the contact point $v_{c,x}$ with respect to the terrain are queried at each timestep prior to evaluating the Bekker-Wong model and can thus be treated as known inputs into the equations. The DARTS implementation accounts for the fact that ERNEST's  wheels are torus-shaped and not cylindrical when determining the location of the reference contact point, however the Bekker-Wong model is still applied assuming a constant wheel width $b$. The longitudinal velocity of the center of the wheel is $v_{x}$, and $\omega$ is the angular velocity of the wheel. The normal stress and longitudinal shear stress applied to the wheel due to the terrain interaction at an arbitrary radial location $\theta$ are $\sigma(\theta)$ and $\tau_{x}(\theta)$, respectively. The radial locations at which the wheel enters and exits the terrain are $\theta_{f}$ and $\theta_{r}$. The angle at which normal stress is maximized is denoted $\theta_{m}$. 

The slip ratio $s$ is calculated as:

\begin{equation}
s = 
\begin{cases}
\frac{v_{c,x}}{v_{c,x} - v_{x}} (1 - e^{-\epsilon_{slip} v_{x}^{2}}) & \text{(} v_{c,x}v_{x} \leq 0 \text{; driving)} \\
\frac{-v_{c,x}}{v_{x}} (1 - e^{-\epsilon_{slip} v_{x}^{2}}) & \text{(} v_{c,x}v_{x} > 0 \text{; braking)} \\
\end{cases},
\end{equation}
where $s$ is clamped between -1 and 1. This is a modification to the approach in~\cite{ishigami2007wheel} because the distinction between braking and driving is determined by comparing the directions of $v_{c,x}$ and $v_{x}$ instead of comparing the magnitudes of $r\omega$ and $v_{x}$. Braking is taken to be whenever the contact velocity is in the same direction as the longitudinal wheel velocity because the shear stress developed at the contact point will oppose the wheel's current motion. This formulation holds even in scenarios such as a wheel translating forward while maintaining a large negative rotation rate. As noted previously, such scenarios could occur when the rover's gimbal system moves the wheel assembly, or when the vehicle to slides down a slope. DARTS also adds a regularization term where $\epsilon_{slip}$ is a positive constant chosen such that $s$ scales smoothly to 0 as longitudinal motion approaches zero ($v_{x} \rightarrow 0$). The slip ratio is ill-defined and discontinuous in this regime. When implemented without $\epsilon_{slip}$ regularization the value of $s$ tends to rapidly thrash between $\pm1$ near longitudinal rest.


The entry angle $\theta_{f}$ can be obtained directly from geometry as:

\begin{equation}
    \theta_{f} = \cos^{-1}{(1 - h/r)}.
\end{equation}
The location of max normal stress and the exit angle can be estimated as:

\begin{equation}
    \theta_{m} = t_{rest}(a_{0} + a_{1}s)\theta_{f},
\end{equation}

\begin{equation}
\theta_{r} = t_{rest}(b_{0} + b_{1}s)\theta_{f} + (t_{rest} - 1) \theta_{f},
\end{equation}

\begin{equation}
    t_{rest} = \min\left(\frac{v_{x}^{2} + v_{c,x}^{2}}{\epsilon_{motion}^{2}}, 1\right),
\end{equation}
where $a_{0}$, $a_{1}$, $b_{0}$, $b_{1}$ are parameters obtained by fitting observations from experimental data. DARTS adds the regularization term $t_{rest}$ to smoothly scale $\theta_{r}$ to $-\theta_{f}$ and $\theta_{m}$ to $0$ as the wheel approaches full rest ($v_{x} \rightarrow 0$ and $v_{c,x} \rightarrow 0$). This ensures the stress distributions are symmetric while the wheel is at rest, preventing net longitudinal forces from being applied.  

The Bekker-Wong model includes a pressure-sinkage formulation to predict normal stresses developed at the wheel's contact patch based on static sinkage into the soil: 

\begin{equation}
    \sigma(\theta) =  r^{n}\left(\frac{k_{c}}{b} + k_{\phi}\right)(\cos{\theta'} - \cos{\theta_{f}})^{n},
\end{equation}

\begin{equation}
\theta' = 
\begin{cases}
\theta & (\theta_{m} \leq \theta \leq \theta_{f}) \\
\theta_{f} - \frac{\theta - \theta_{r}}{\theta_{m} - \theta_{r}}(\theta_{f} - \theta_{m}) & (\theta_{r} \leq \theta < \theta_{m})
\end{cases},
\end{equation}
where $k_{c}$ is the cohesion modulus, $k_{\phi}$ is the friction modulus, and $n$ is the sinkage exponent. 

The model also includes Janosi and Hanamoto's traction-slip relation to predict shear stresses developed by a driving or braking wheel:

\begin{equation}
    \tau_{x}(\theta) = (c + \sigma(\theta)\tan\phi)(1 - e^{-|j_{x}(\theta)|/k_{x}}),
    \label{eq:tau_x}
\end{equation}

\begin{equation}
    \tau_{y}(\theta) = (c + \sigma(\theta)\tan\phi)(1 - e^{-|j_{y}(\theta)|/k_{y}}) + \frac{F_{s}(\theta)}{rb},
    \label{eq:tau_y}
\end{equation}
where $c$ is the soil cohesion, $\phi$ is the soil internal friction angle, and $k_{x}$, $k_{y}$ are the shear deformation moduli in the x and y directions, respectively. The soil deformations, $j_{x}(\theta)$ and $j_{y}(\theta)$ are estimated via:

\begin{equation}
    j_{x}(\theta) = r(\theta_{f} - \theta - (1 - s)(\sin{\theta_{f}} - \sin{\theta})),
\end{equation}

\begin{equation}
    j_{y}(\theta) = r(1 - s)(\theta_{f} - \theta)\tan{\beta},
\end{equation}
where $\beta = \tan^{-1}(v_{y} / |v_{x}|)$ is the sideslip angle. An approximate lateral bulldozing force $F_{s}(\theta)$ is also applied based on the formulation proposed by Hegedus~\cite{hegedus1960bulldozing} to account for the action of the wheel's sidewalls pushing through the soil:

\begin{equation}
    F_{s}(\theta) = R_{b}(\theta) \cdot \left(r - h(\theta)\cos{\theta}\right),
\end{equation}

\begin{equation}
    R_{b}(\theta) = D_{1}\left(c h(\theta) + D_{2}\frac{\rho h^{2}(\theta)}{2}\right),
\end{equation}

\begin{equation}
    D_{1} = \cot{X_{c}} + \tan(X_{c} + \phi),
\end{equation}

\begin{equation}
    D_{2} = \cot{X_{c}} + \cot^{2}{X_{c}} / \cot{\phi},
\end{equation}

\begin{equation}
    X_{c} = \frac{\pi}{4} - \frac{\phi}{2},
\end{equation}



\noindent where $\rho$ is the soil density. In the DARTS implementation the penetration depth $h(\theta)$ is estimated by linearly interpolating $h$ as a function of $\theta$ to be zero at $\theta_{f}$ and $\theta_{r}$.

In DARTS the absolute value of $j_{x}(\theta)$ and $j_{y}(\theta)$ are taken in Eq.~\ref{eq:tau_x}-\ref{eq:tau_y} because the shear stresses are updated as follows:

\begin{equation}
    \tau_{x}(\theta) \rightarrow \text{sign}( -{v_{c,x}}) \cdot \min\left( \frac{|v_{c,x}|}{\epsilon_{contact,x}}, 1 \right) \cdot \tau_{x}(\theta),
\end{equation}

\begin{equation}
    \tau_{y}(\theta) \rightarrow \text{sign}( -{v_{c,y}}) \cdot \min\left( \frac{|v_{c,y}|}{\epsilon_{contact,y}}, 1 \right) \cdot \tau_{y}(\theta),
\end{equation}
where the $\sign(\cdot)$ terms apply the correct sign to the final shear stress values according to the requirement that shear stress should always oppose the terrain-relative motion of the wheel at the contact point. The $\min(\cdot)$ terms smoothly scale the shear stresses to zero when the contact velocity magnitude goes below a positive regularization constant $\epsilon_{contact}$. 

Finally, numerically integrating the stress distributions across the contact patch yields the overall normal and tractive forces applied to each wheel:  

\begin{equation}
    F_{x} = rb\int_{\theta_{r}}^{\theta_{f}}(\tau_{x}(\theta)\cos{\theta} - \sign(v_{x})\sigma(\theta)\sin{\theta})d\theta,
\end{equation}

\begin{equation}
    F_{y} = rb\int_{\theta_{r}}^{\theta_{f}}(\tau_{y}(\theta))d\theta,
\end{equation}

\begin{equation}
    F_{z} = rb\int_{\theta_{r}}^{\theta_{f}}(\sign(v_{x})\tau_{x}(\theta)\sin{\theta} + \sigma(\theta)\cos{\theta})d\theta.
\end{equation}

Forming $\textbf{F} = [F_{x}, F_{y}, F_{z}]$ provides the resultant force vector applied to the wheel at the center of the contact patch. The $\sign(v_{x})$ terms correct for the fact that the integration direction of $\theta$ will reverse depending on the direction of longitudinal travel. With this correction, the integrals can be set up as though the wheel were moving forward (moving to the right in Fig~\ref{fig:bekker_diagram}) regardless of if the wheel is traveling backwards.

The DARTS implementation of the Bekker-Wong model does not keep track of prior terrain deformation, unlike, for example, the Soil Contact Model (SCM) implementation in Project Chrono~\cite{tasora2018chronoscm}. Subsequent wheels passing over the same terrain patch will see the same undeformed terrain as the lead wheel. Further, the Bekker-Wong model does not account for soil dislocation or excavation. One can readily identify scenarios in which the absence of terrain deformation or dislocation simplifies the learning problem for the RL algorithm. For example, the rover cannot dig itself into a rut. Such terrain geometries could, however, be explicitly included in the training set. Conversely, there are scenarios in which the lack of terrain deformation increases task difficulty for the RL algorithm. For example, the rover cannot flatten terrain features, and is therefore required to adapt to the given terrain geometry. This modeling limitation was identified as a significant sim-to-real gap early on in the the study.

The terramechanics implementation is intentionally kept simple for the present work. The governing equations can be evaluated at speeds exceeding real time, which is essential for RL as large volumes of training data must be generated efficiently. Policies trained in simulation using the Bekker–Wong model exhibited consistent and physically plausible behaviors, transferring effectively to real-world experiments, including on loose sand. On this basis, large-scale training with the Bekker–Wong model was deemed sufficient for the present work. Future work should incorporate a subset of training using higher-fidelity terramechanics models that capture soil transport and terrain deformation, with the aim of complementing and refining the behaviors learned under the lower-fidelity Bekker–Wong formulation.

A complete list of the terramechanics parameters used in simulation is given in Table~\ref{tab:terra-params}.

\begin{table}[h!]
\centering
\caption{Bekker-Wong parameter values used in simulation}
\label{tab:terra-params}
\resizebox{\columnwidth}{!}{
\begin{tabular}{l|l|p{1.2cm}|p{1.5cm}}
    \hline
    \textbf{Parameter} & \textbf{Description} & \textbf{Dry sand} \cite{jia2013robotica} & \textbf{M90 Mars simulant} \cite{oravec2021geotechnical} \\
    \hline
    $\rho$ ($\frac{kg}{m^{3}}$) & Soil density & 1600 & 1520 \\
    $c$ ($Pa$) & Cohesion & $1200^{\dagger}$ & 2000 \\
    $\phi$ ($deg$) & Internal friction angle & $33.3^{\dagger}$ & 35.0 \\
    $n$ (-) & Sinkage exponent & $1.9^{\dagger}$ & 1.3 \\
    $k_{c}$ ($\frac{kN}{m^{n+1}}$)& Cohesion modulus & 10.3 & 572.1 \\
    $k_{\phi}$ ($\frac{kN}{m^{n+2}}$)& Friction modulus & 5309.4 & 4915.3 \\
    $K$ ($m$) & Shear modulus & 0.015 & 0.0254 \\
    $a_{0}$, $a_{1}$ (-) & Coefficients for $\theta_{m}$ & 0.43, 0.32 & 0.43$^{\S}$, 0.32$^{\S}$ \\
    $b_{0}$, $b_{1}$ (-) & Coefficients for $\theta_{r}$ & -0.16, 0.0 & -0.16$^{\S}$, 0.0$^{\S}$ \\
    $\epsilon_{slip}$ $\left(\frac{m}{s}\right)^{-2}$ & Slip ratio regularization$^{*}$ & 100 & 100 \\
    $\epsilon_{motion}$ $\left(\frac{m}{s}\right)$ & Motion threshold$^{*}$ & $5\cdot 10^{-5}$ & $5\cdot 10^{-5}$ \\
    $\epsilon_{contact}$ $\left(\frac{m}{s}\right)$ & Contact velocity threshold$^{*}$ & 0.01 & 0.01 \\
    \hline
    \multicolumn{4}{l}{$^{\dagger}$manually adjusted to match drawbar pull experiments (see Fig~\ref{fig:drawbar_pull_comp})} \\
    \multicolumn{4}{l}{$^{*}$values chosen provide reasonable integrator performance} \\
    \multicolumn{4}{l}{$^{\S}$values taken to be same as dry sand due to lack of data}
\end{tabular}
}
\end{table}

\begin{figure}
	\centering
	\includegraphics[width=0.5\textwidth]{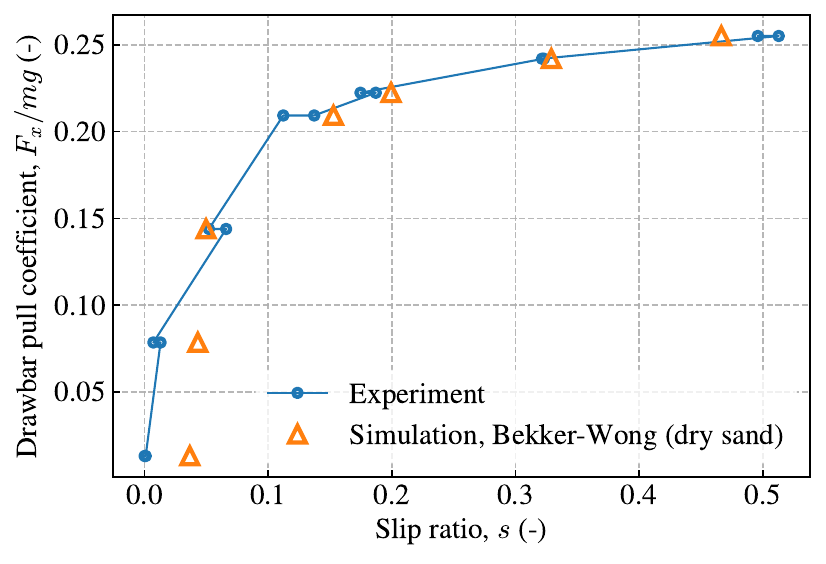}
	\caption{Drawbar pull comparison between simulation and experiment. The DARTS simulation uses manually-adjusted parameters for dry sand. The experimental values are averaged over three trials.}
	\label{fig:drawbar_pull_comp}
\end{figure}

We adapt Bekker-Wong terramechanics parameters from existing literature in this work. We employed a two-phase approach to parameter selection. First, to verify the Bekker-Wong model was correctly implemented, we manually tuned dry sand values from~\cite{jia2013robotica} to fit experimental drawbar pull data collected in a preliminary testing area of dry, hard-packed sand (Fig~\ref{fig:drawbar_pull_comp}). 

The results shown in Fig~\ref{fig:drawbar_pull_comp} demonstrate that the model produces physically accurate results when provided with tuned values. Notably, increasing the sinkage exponent $n$ from 1.1 to 1.9 results in a much closer fit, which we hypothesize is due to the fact that the Bekker-Wong model is formulated assuming a rigid wheel when in reality ERNEST's wheels are highly flexible. Next, for the majority of the simulation training, we transitioned to using M90 Mars sand simulant values from~\cite{oravec2021geotechnical} to match the primary testing environment. We chose not to tune the M90 parameters to fit experimental data. As mentioned previously, the primary objective of the simulator is to produce physically plausible agent behaviors that transfer to real-world scenarios. Once the model was validated, we decided it was acceptable to train with unadjusted parameters.

\subsection{RIGID CONTACT MODEL}

\begin{figure}
	\centering
	\includegraphics[width=0.2\textwidth]{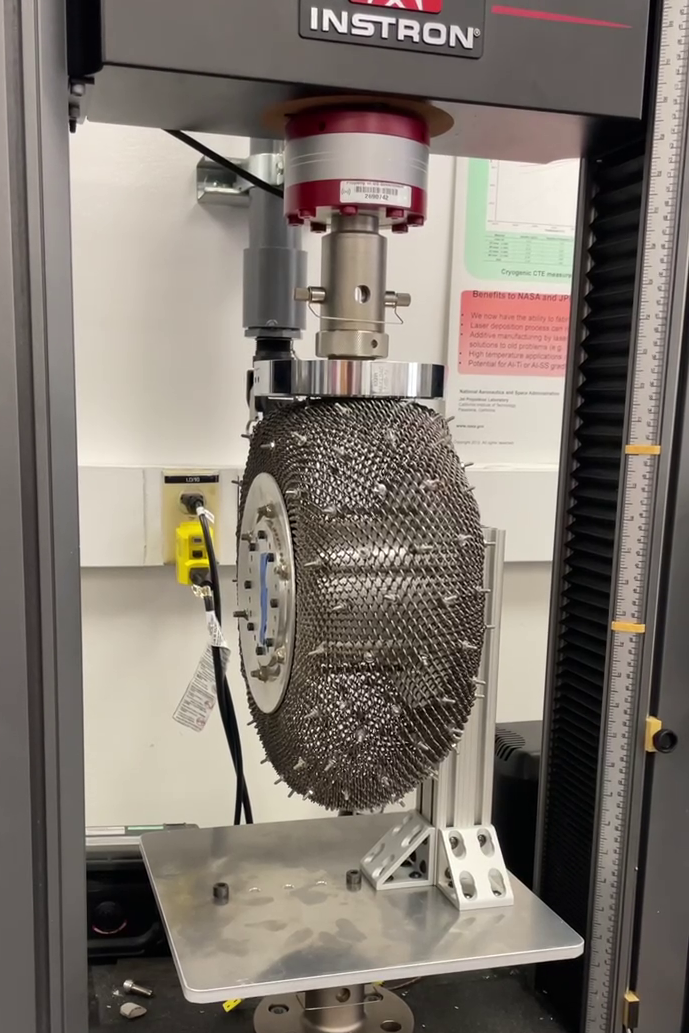}
	\caption{Wheel compression experiment used to determined the stiffness of the wheel interactions with rigid surfaces.}
	\label{fig:wheel_compression_test}
\end{figure}

A linear spring–damper contact model with Coulomb friction is used to represent contact with rigid surfaces, such as rocks or steps. Since the friction coefficient depends on the properties of both contacting surfaces, it is randomized during training to account for variability in real-world conditions. In contrast, the compliance of the wheels is assumed to dominate that of any encountered rigid obstacle, therefore the stiffness and damping parameters are identified \textit{a priori}. The stiffness coefficient is determined as \qty{20}{\kilo\newton\per\meter} from wheel compression experiments (see Fig.~\ref{fig:wheel_compression_test}), whereas the damping coefficient is estimated as \qty{20}{\kilogram\per\second} based on the rebound height observed in wheel drop tests.

\section{REINFORCEMENT LEARNING FRAMEWORK}
\label{sec:rl_framework}

\subsection{TRAINING ALGORITHM}

Conventional planning approaches are challenged by the nature of the continuous wheel-terrain contact and the complexity of terramechanics interactions when attempting to plan a physically viable path. Even when using simplifying assumptions about the physics and continuity, at the expense of the solution accuracy, the computation time required for planning jeopardizes the ability to react to deviations from the expected trajectory, whether caused by perception inaccuracies or alterations of the environment. Therefore, we employ reinforcement learning to develop a highly responsive controller capable of rapidly determining the continuous control law to apply to the Active Gimbal Suspension based on the challenges encountered by the rover in real time. This controller, which ultimately consists of a unique, small neural network that intrinsically contains all the "planning intelligence" acquired during offline training, provides control setpoints with very high computational efficiency.

To train the neural network controller, we use\footnote{Our implementation of TD3 can be found at \url{https://github.com/arthur-bouton/MachineLearning}.} the Twin Delayed Deep Deterministic Policy Gradient algorithm (TD3)~\cite{fujimoto2018addressing}, which is itself based on the Deep Deterministic Policy Gradient algorithm (DDPG)~\cite{lillicrap2015continuous}, augmented with several heuristic techniques designed to improve training stability.
Unlike the widely adopted Proximal Policy Optimization (PPO)~\cite{schulman2017proximal}, this algorithm is off-policy, i.e. the training experience does not need to be generated by the current policy being optimized, which is important here for three reasons.

First, it allows the collected experience to be reused multiple times while the neural networks slowly converge. This substantially improves data efficiency, which is particularly desirable when simulations are computationally expensive due to complex physics such as terramechanics.

Second, it provides explicit control over the exploration process, i.e. it allows us to determine when to deviate from the current policy by executing a random action, and to choose the associated probability distribution. Consequently, the state–action space can be explored more efficiently, enabling faster discovery of successful behaviors and better avoidance of local optima compared with purely Gaussian exploration.

Finally, the off-policy nature of the algorithm makes it possible to aggregate experience generated in independent training instances and subsequently train a new controller from this pooled dataset. This approach ultimately enables the integration of multiple capabilities into a single unified neural network. Indeed, attempting to directly train a neural network policy to solve all obstacle scenarios simultaneously proved intractable, likely due to reduced network plasticity under prolonged concurrent exploration and policy optimization~\cite{lyle2023understanding}. Instead, we first train several independent pairs of actor and critic networks, each specialized to solve a specific class of challenges. We then aggregate the most recent million samples of experience produced by each pair and train a newly initialized actor–critic pair solely on this combined dataset. Because this second phase does not require additional simulation rollouts, it is significantly faster. The resulting policy effectively consolidates the capabilities progressively acquired by the specialized networks across the different training scenarios.

In both training phases, the actor and critic networks each comprise two hidden layers of 512 fully connected neurons with Rectified Linear Unit (ReLU) activation functions. The actor network takes the state as input to the first layer and outputs the actions through a final layer of three hyperbolic tangent (tanh) neurons that map the outputs to the desired control range. The critic network takes the concatenated state–action vector as input to the first layer, while the action vector is also reintroduced at the second layer. It terminates with a single linear neuron that provides the state–action value estimate.

\subsection{OBSERVATIONS AND ACTIONS}

\begin{figure}
	\centering
	\includegraphics[width=0.5\textwidth]{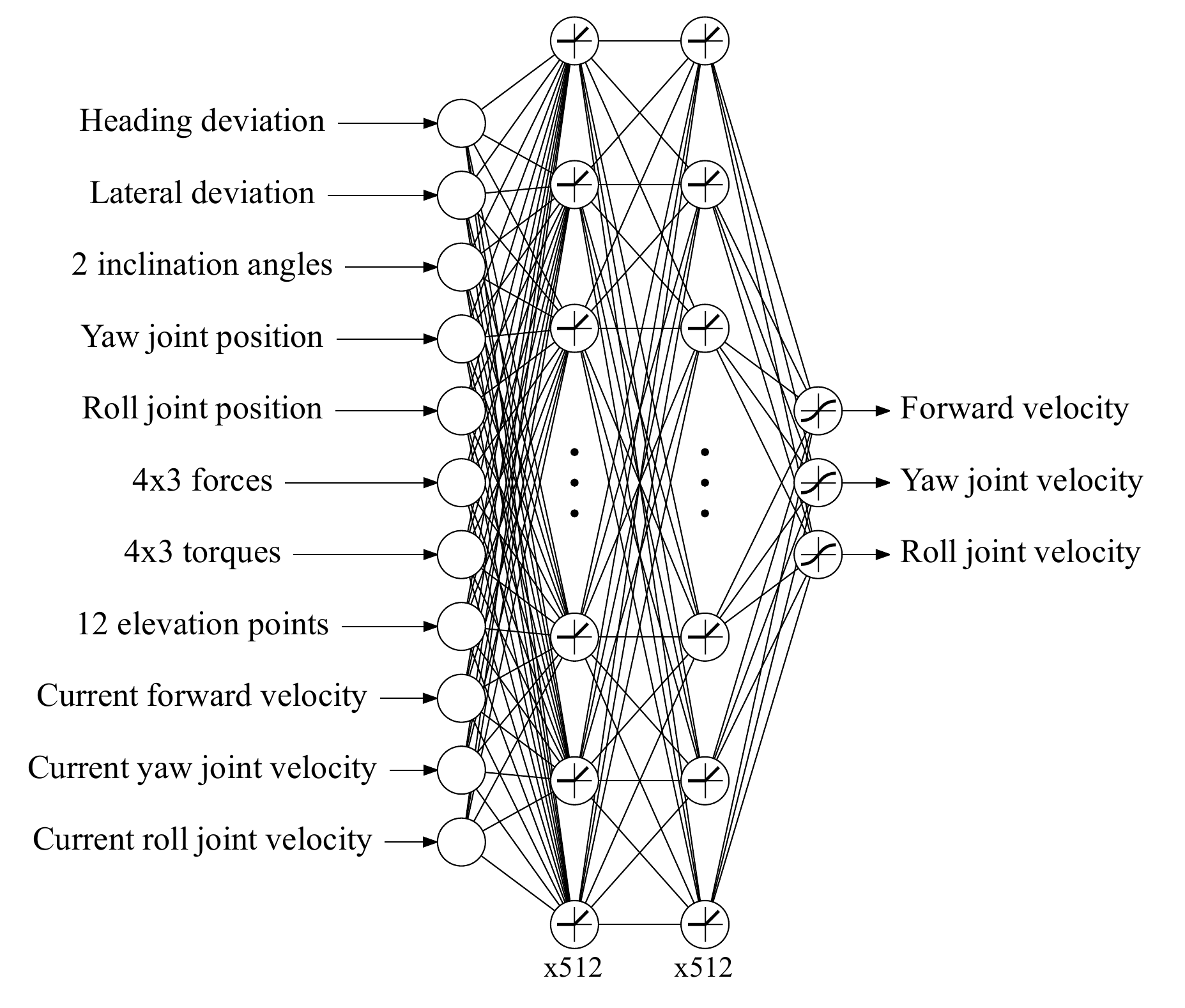}
	\caption{Diagram of the actor neural network, along with its inputs and outputs.}
	\label{fig:actor_network}
\end{figure}

The architecture of the actor network, along with its inputs and outputs, is illustrated in Fig.~\ref{fig:actor_network}. All the inputs are normalized with respect to their maximum expected values before being fed to the neural network. The integration of the neural network within the overall perception and control architecture is summarized in Fig.~\ref{fig:system_diagram}.

The action vector comprises the rover forward velocity and the angular velocities of the yaw and roll joints of the Active Gimbal Suspension. These three components correspond respectively to the variables $V_x$, $\dot{\psi}$, and $\dot{\phi}$ used in Section~\ref{sec:wheel_synchronization}, in particular in Eq.~\ref{eq:y}, to determine the required wheel rotational speeds. We use velocity commands as actions because they provide a more immediate interface with the system dynamics, reducing the effective order and nonlinearity seen by the policy and thus facilitating learning. Moreover, velocity limits map naturally to actuator constraints, enabling straightforward enforcement of safety bounds.

Including the forward velocity as an action enables the policy to modulate the rover speed during maneuvers, e.g., slowing down when negotiating tight turns to allow the yaw joint of the gimbal to steer the wheels without overshooting the trajectory. However, to prevent reverse motion while the perception system is forward-facing, this velocity is restricted to the range \qtyrange{0}{0.2}{\meter\per\second}. The yaw and roll joint velocities are bounded by \qty{\pm 20}{\degree\per\second} and \qty{\pm 10}{\degree\per\second}, respectively. Although ERNEST is capable of higher speeds, these limits are conservatively set for testing safety.

The observations used by the rover to represent its current state and determine the next action consist of:
\begin{itemize}
\item Heading deviation: the yaw angle between the current orientation of the chassis and the direction of the desired path.
\item Lateral deviation: the signed distance of the chassis from the desired path.
\item Inclination angles: roll and pitch angles of the chassis with respect to gravity.
\item Gimbal joint positions: positions of the yaw and roll joints of the Active Gimbal Suspension.
\item Force--torque measurements: six-axis, pre-processed signals from the four force--torque sensors located above the wheel assemblies.
\item Elevation sampling: terrain elevations measured relative to the chassis at 12 points located in front of and beneath the rover.
\item Current action: the most recently executed action vector, i.e., the currently applied control inputs.
\end{itemize}

Before being used as inputs to the neural networks, the measurements from the force–torque sensors (FTS) undergo three processing steps. First, the data from each of the six axes, sampled at \qty{400}{\hertz}, are filtered using a first-order recursive low-pass filter with a cutoff frequency of \qty{0.5}{\hertz}. Then, when required by the RL controller (i.e., at \qty{2}{\hertz}), calibration biases are subtracted from each axis. Finally, the average value of each axis across the four wheel FTS is also subtracted.

Force–torque sensors are prone to measurement drift over time, notably due to temperature variations. This final step therefore mitigates the impact of correlated drifts by focusing on relative differences between wheel FTS measurements rather than on their absolute values. As a result, the bias calibration is primarily needed to ensure consistent offsets between sensors rather than to recover the true absolute force–torque values.

The calibration procedure consists of recording the filtered FTS readings when all sensors are expected to measure the same load. In practice, this is achieved by supporting the rover’s chassis with a jack so that the four identical wheel assemblies hang freely, and then storing the resulting measurements as bias values. The jack is used for speed and convenience, although the procedure could also be performed autonomously by ERNEST, since its active gimbal suspension allows it to lift each wheel off the ground sequentially.

Force–torque sensors are also inherently noisy. Furthermore, these measurements are more likely to deviate from their simulated counterparts than the other state inputs, due to the difficulty of accurately modeling the complex physical interactions between the wheels and the ground. To mitigate over-reliance of the policy on precise FTS readings, Gaussian noise is injected into the corresponding components of both the state and next-state vectors during training. This noise is resampled each time a transition is drawn, so that a single stored transition can be reused multiple times with different noise realizations. The injected noise follows a zero-mean Gaussian distribution with a standard deviation of \qty{10}{\newton} for each force component and \qty{2}{\newton\meter} for each torque component.

\begin{figure}
	\centering
	\newcommand{\widthratio}{0.4}

	\renewcommand{\thesubfigure}{a}%
	\subfloat[]{
        \includegraphics[width=\widthratio\textwidth]{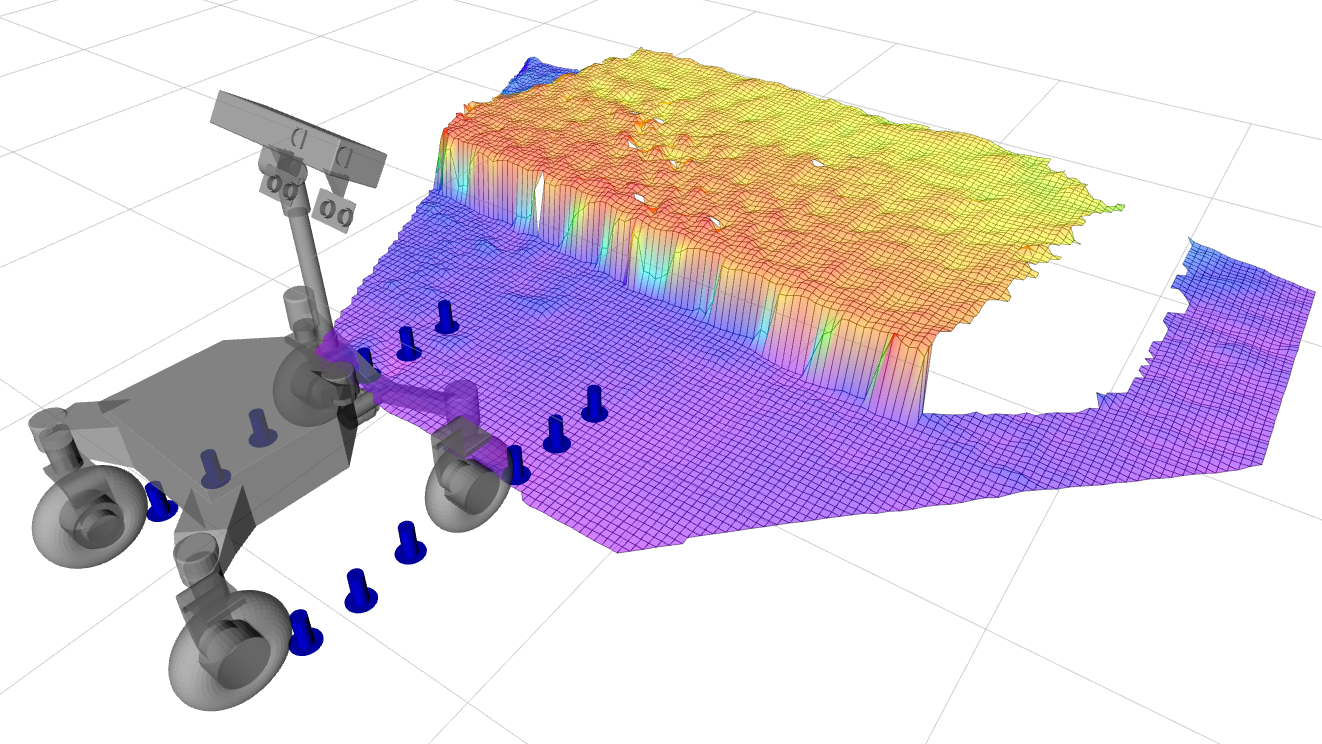}
        \label{fig:elevation_map_step}
    }
    \hfil
    \renewcommand{\thesubfigure}{b}%
	\subfloat[]{
        \includegraphics[width=\widthratio\textwidth]{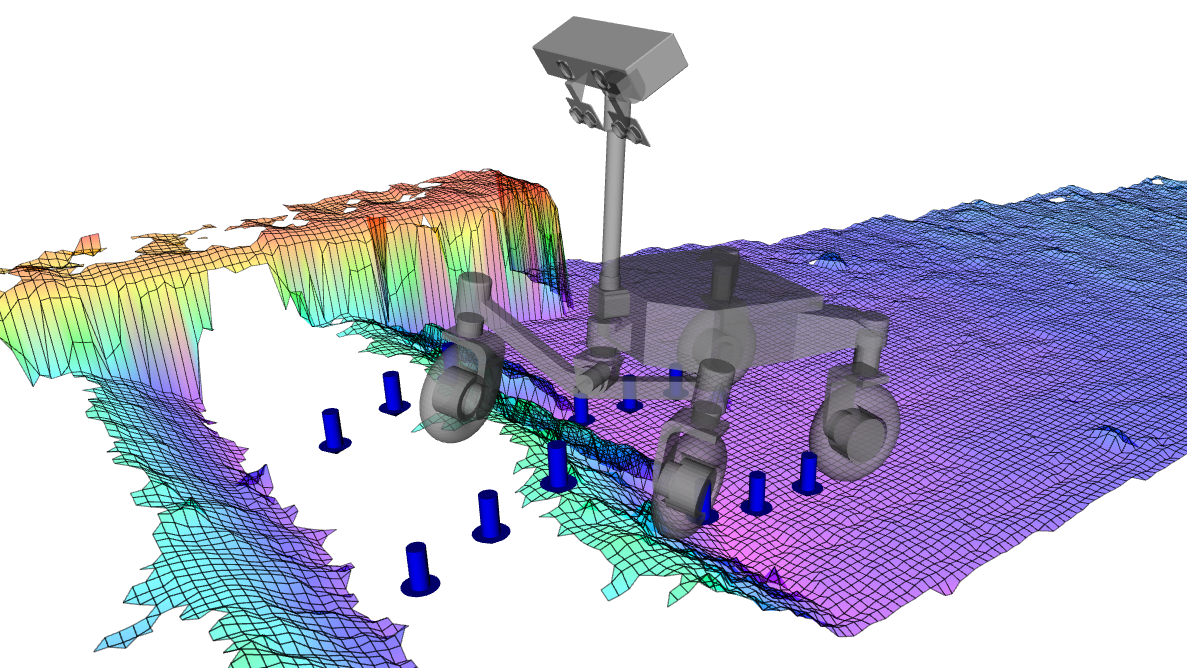}
        \label{fig:elevation_map_ripples}
    }
    \hfil
    \renewcommand{\thesubfigure}{c}%
	\subfloat[]{
        \includegraphics[width=\widthratio\textwidth]{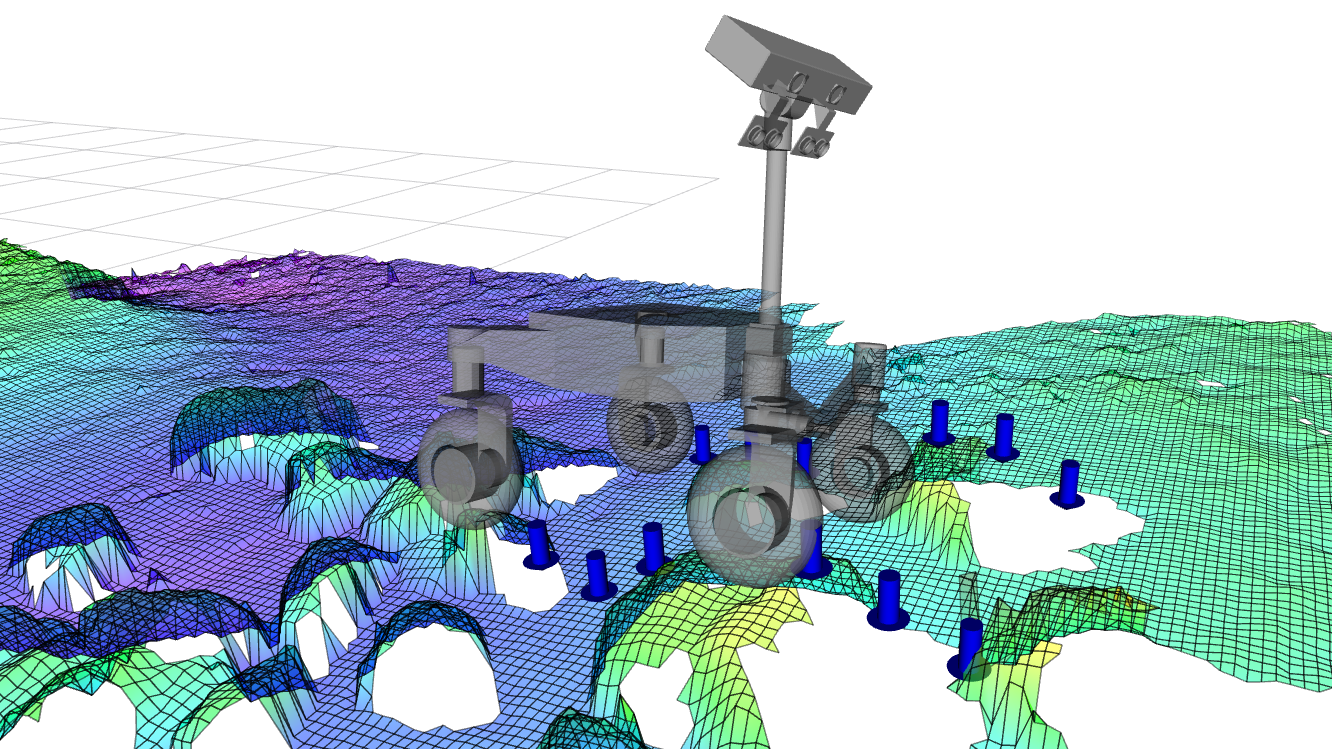}
        \label{fig:elevation_map_rocks}
    }

	\caption{Onboard elevation maps produced by stereo vision.
    \protect\subref{fig:elevation_map_step} Elevation map at initialization, before rover motion, in front of a step, illustrating the field of view of the stereo-vision mapping system.
    \protect\subref{fig:elevation_map_ripples} Elevation map while approaching sand ripples.
    \protect\subref{fig:elevation_map_rocks} Elevation map in a rock field.
    The blue downward arrows show the (x,y) coordinates at which the terrain elevation relative to the chassis is sampled to compose the observed state of the rover. When a sampling point falls within a map hole caused by occlusion, the elevation is estimated through linear interpolation.}
	\label{fig:elevation_sampling}
\end{figure}

The state vector also includes 12 samples of terrain elevation in front of the wheels. The elevation values are obtained from the elevation map constructed using stereo vision. They are expressed relative to the current chassis position and orientation, such that when the rover is resting with all four wheels on perfectly flat terrain, all sampled values are zero, regardless of the absolute ground altitude or slope. The sampling points are arranged in groups of three in front of each wheel, aligned along the forward direction, and spaced \qty{20}{\centi\meter} apart, as shown in Fig.~\ref{fig:elevation_sampling}. This sparse sampling strategy is chosen to promote generalization. The coordinates of the sampling points are defined in the chassis frame. Consequently, the points located in front of the front wheels do not follow the motion of the bogie, as illustrated in Fig.~\ref{fig:elevation_map_ripples}. When a sampling point falls within a region of missing data in the elevation map (e.g., due to occlusions), the elevation value is estimated by linear interpolation along the sampling direction, as shown in Fig.~\ref{fig:elevation_map_ripples} and Fig.~\ref{fig:elevation_map_rocks}.

\begin{figure}
	\centering
	\includegraphics[width=0.5\textwidth]{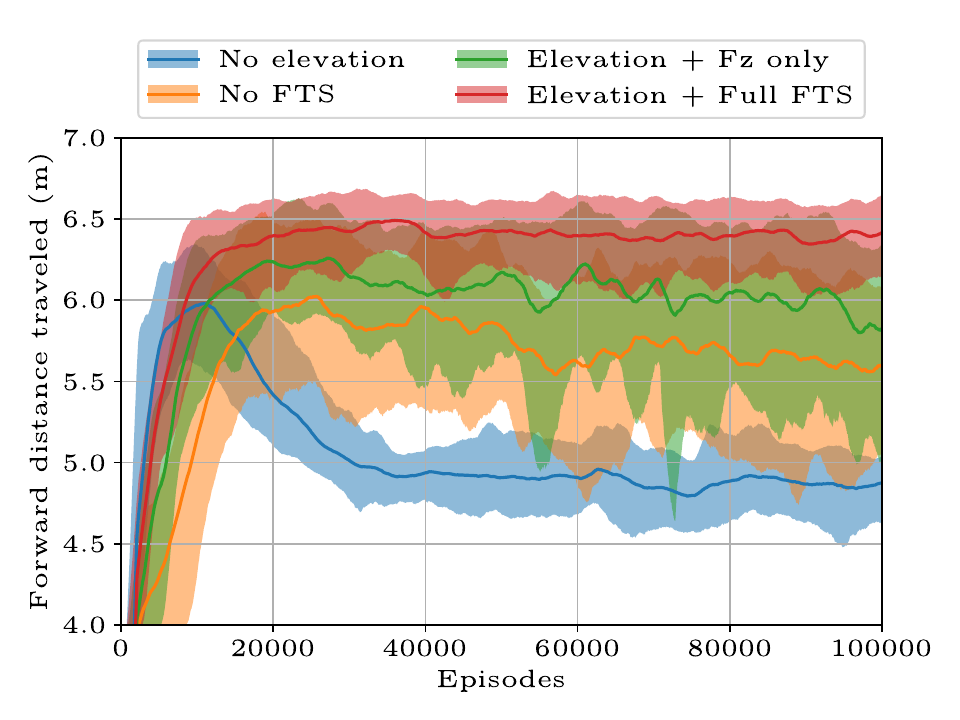}
	\caption{Comparison of the evolution of performance during training on the same gradually increasing slope to \qty{25}{\degree} over 10 training instances for each different set of observation inputs. The shaded regions delineate the minimum and maximum performance among the 10 training instances.}
	\label{fig:state_comp}
\end{figure}

We evaluated whether the policy could perform effectively without elevation inputs, without force–torque sensor (FTS) inputs, or using only vertical force measurements combined with elevation sampling. The results show that combining the full FTS information with elevation inputs yields significantly better performance. Fig.~\ref{fig:state_comp} presents a performance comparison across multiple training instances on the same gradually increasing slope, where the complete set of inputs is required to achieve an optimal crawling gait that enables further ascent.

When evaluating the controller on the physical rover without including the current action in the state, we observed that the system was prone to becoming trapped in limit cycles when interacting with soft soil. To mitigate this behavior, the previous actions were incorporated into the state representation. We investigated the inclusion of longer action histories by augmenting the state with multiple past actions, but found that using only the most recent action, corresponding to the currently applied control inputs, was sufficient to significantly improve performance while preserving generalization.

\subsection{REWARD}

The reward associated with each simulated transition shapes the critic network through dynamic programming, such that it ultimately estimates the expected cumulative discounted return for any state--action pair. In turn, the policy is updated using the gradient of the critic with respect to the action so as to maximize this return. The reward function therefore directly governs the learned behavior. In this work, it comprises three terms: (i) a term encouraging forward progress along the desired path while penalizing regression, (ii) a term penalizing lateral deviation from this path, and (iii) a term penalizing the roll joint torque of the Active Gimbal Suspension to limit unnecessary energy consumption.

Given the desired path heading angle $\theta_d$, and $(x_0, y_0)$ the coordinates of a point lying on this path, the coefficients of the corresponding line can be written as
\begin{align}
    a &= -\sin(\theta_d), \\
    b &= \cos(\theta_d), \\
    c &= -a\,x_0 - b\,y_0.
\end{align}
The progress along the path between two consecutive time steps is then computed as
\begin{equation}
    \Delta s = b\bigl(x_{\mathrm{chassis}, t} - x_{\mathrm{chassis}, t-1}\bigr)
    - a\bigl(y_{\mathrm{chassis}, t} - y_{\mathrm{chassis}, t-1}\bigr),
\end{equation}
where $(x_{\mathrm{chassis}, t},\, y_{\mathrm{chassis}, t})$ denote the horizontal position of the chassis at time $t$.
The lateral deviation from the path is given by
\begin{equation}
    \delta = a\,x_{\mathrm{chassis}, t} + b\,y_{\mathrm{chassis}, t} + c.
\end{equation}
The reward function is finally defined as
\begin{equation}
    R = w_s\,\frac{\Delta s}{\Delta t}
    - w_\delta\,\delta^2
    - w_\tau\,\lvert \tau_{\phi} \rvert,
\end{equation}
where $w_s = 2$, $w_\delta = 0.1$, and $w_\tau = 10^{-4}$ are tunable weights, $\Delta t = \qty{0.5}{\second}$ corresponds to the control period of the reinforcement learning policy, and $\tau_{\phi}$ is the torque in the roll joint of the active gimbal suspension.

\subsection{EXPLORATION STRATEGY}

\begin{figure}
	\centering
	\includegraphics[width=0.4\textwidth]{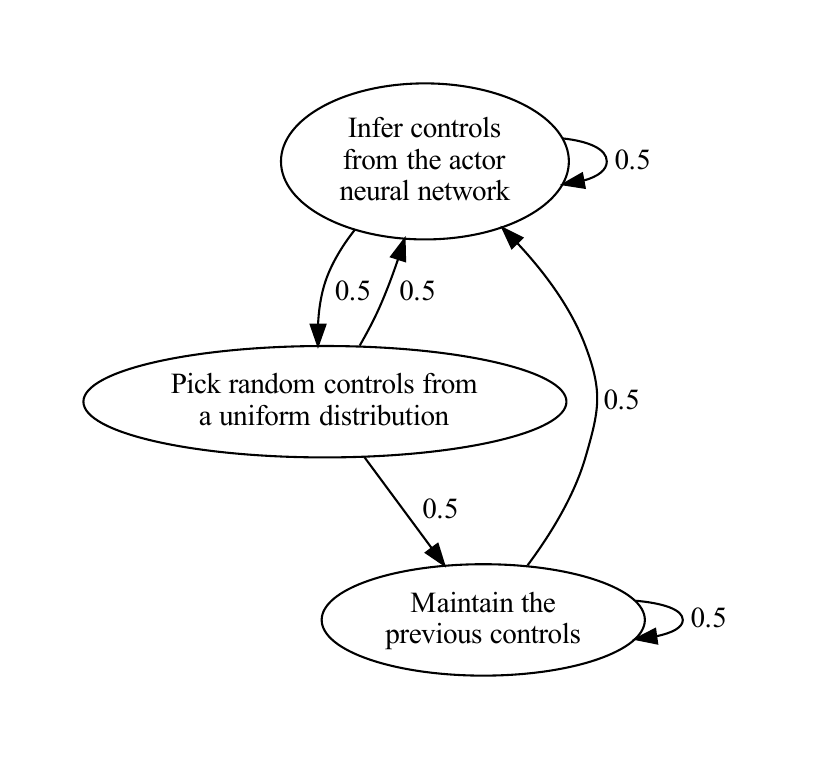}
	\caption{Custom \textepsilon-greedy exploration strategy used during training, represented as a Markov process. Numbers indicate the probability of each transition at each control timestep.}
	\label{fig:exploration_graph}
\end{figure}

Because an \emph{off-policy} algorithm is employed, the exploration strategy can be freely defined, i.e., when and from which distribution to sample actions that differ from the policy. In this work, we adopt a custom \textepsilon-greedy exploration strategy, represented as a Markov process in Fig.~\ref{fig:exploration_graph}, which has been found to be the most effective and was first introduced in~\cite{bouton2023marcel}. At each decision step, the controller either follows the policy action or samples a random action with equal probability. Random actions are sampled from a uniform distribution spanning the full admissible control range. When a random action is selected, it is either maintained for the next step or the system reverts to the policy, again with equal probability. This mechanism allows a random action to persist over multiple policy steps, enabling its effects to manifest over longer time horizons. It also improves the exploration of the state space by forcing the system to stray further from the trajectory prescribed by the current policy, compared with the more erratic behavior induced by standard \textepsilon-greedy or Gaussian exploration.

\subsection{TRAINING SCENARIOS}

\begin{figure}
	\centering
	\newcommand{\widthratio}{0.24}

	\renewcommand{\thesubfigure}{a}%
	\subfloat[]{%
        \includegraphics[width=\widthratio\textwidth]{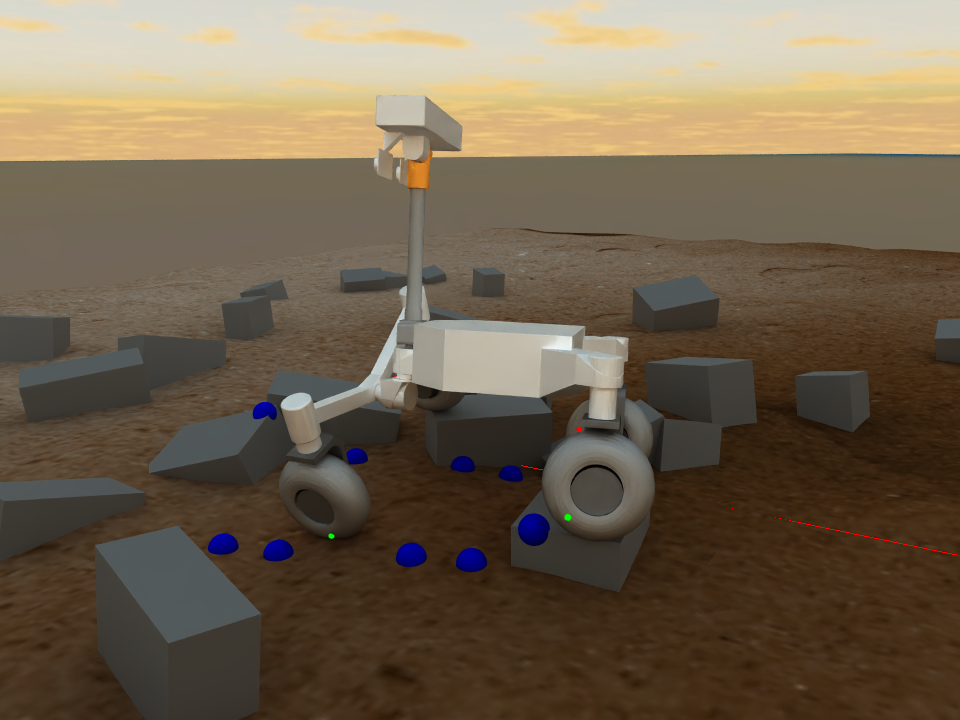}%
        \label{fig:darts_rubble}%
    }
    \hfil
    \renewcommand{\thesubfigure}{b}%
	\subfloat[]{%
        \includegraphics[width=\widthratio\textwidth]{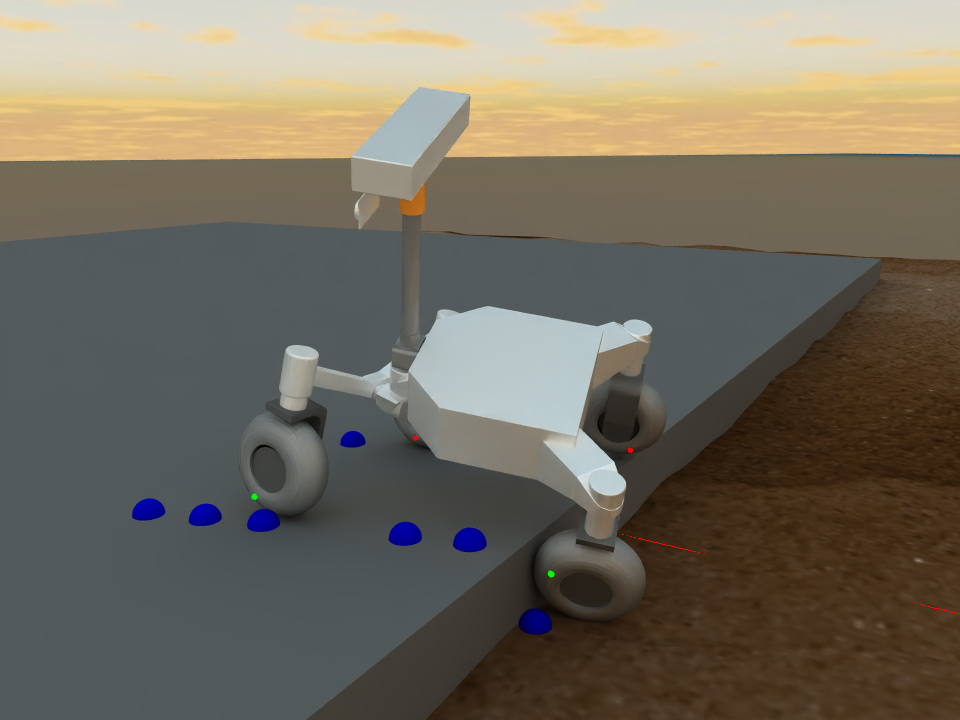}%
        \label{fig:darts_step}%
    }

	\renewcommand{\thesubfigure}{c}%
	\subfloat[]{%
        \includegraphics[width=\widthratio\textwidth]{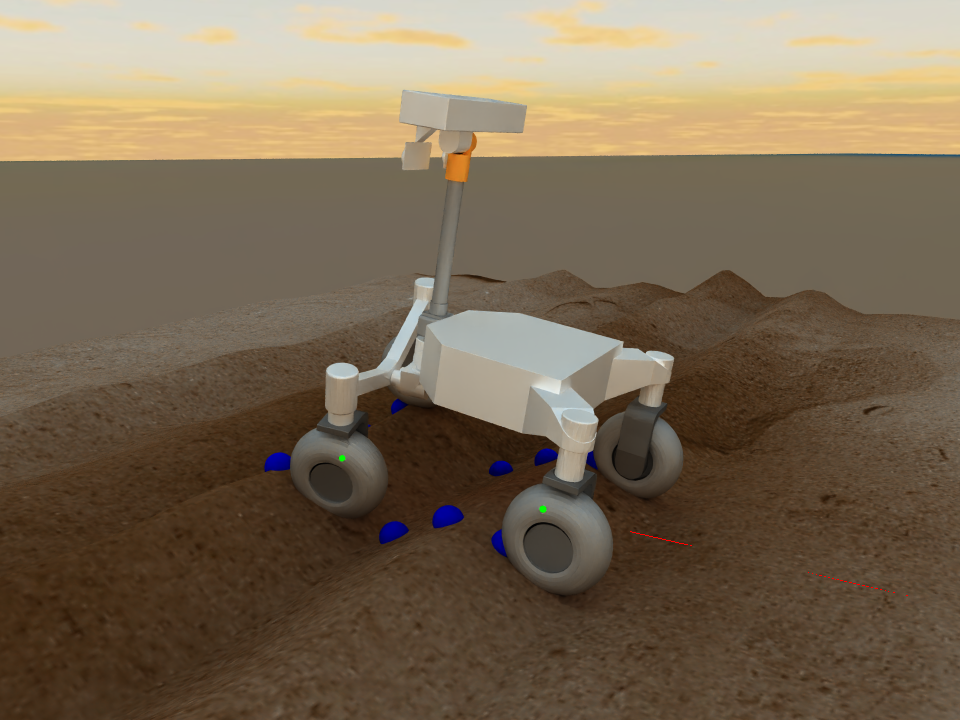}%
        \label{fig:darts_ripples}%
    }
	\hfil
	\renewcommand{\thesubfigure}{d}%
	\subfloat[]{%
        \includegraphics[width=\widthratio\textwidth]{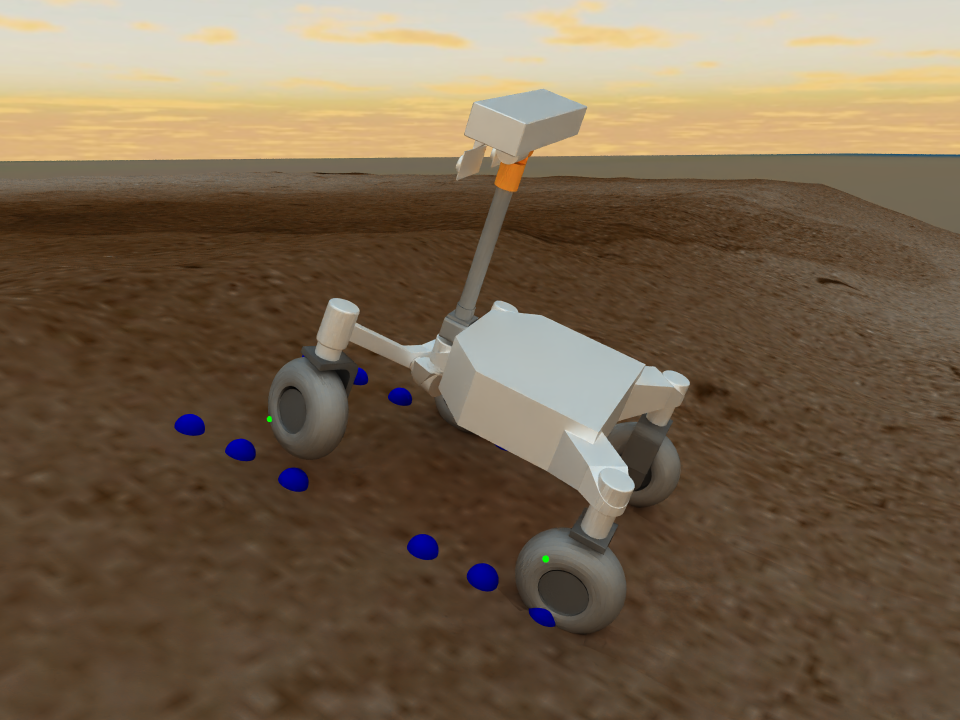}%
        \label{fig:darts_slope}%
    }

	\caption{The four different scenarios used for training using DARTS simulations: \protect\subref{fig:darts_rubble} Rock fields. \protect\subref{fig:darts_step} Step obstacles. \protect\subref{fig:darts_ripples} Ripples. \protect\subref{fig:darts_slope} Slopes. The Blue dots show the elevation sampling points that the neural network currently "sees". The brown heightmap surface simulates soft Bekker-Wong terramechanics, while the gray obstacles simulate hard contacts with Coulomb friction.}
	\label{fig:darts_scenarios}
\end{figure}

Independent policies are first trained on four distinct terrain classes, as illustrated in Fig.~\ref{fig:darts_scenarios}: rock fields, step obstacles, ripples, and slopes.

For the step scenario, each simulation trial runs for \qty{30}{\second} before being reset with a newly generated terrain instance. In the remaining scenarios, a longer trial duration of \qty{60}{\second} is used. If the rover reaches the end of the map, the trial is terminated early and reset.

In all cases, the main terrain surface is defined by a heightmap with terramechanical properties. At the beginning of each trial, two octaves of Perlin noise are generated and superimposed onto the base surface to introduce stochastic irregularities. The corresponding spatial periods are \qty{1}{\meter} and \qty{0.5}{\meter}. The amplitude of the lower-frequency component is defined in the ``Ground'' section of Table~\ref{tab:rand_params}, while that of the higher-frequency component is set to half this value.

The parameters governing domain randomization for each scenario are summarized in Table~\ref{tab:rand_params}.

\begin{table}
    \centering
    \begin{tabular}{lcccc}
        \toprule
        \textbf{Parameter} & \textbf{Rock field} & \textbf{Step} & \textbf{Ripples} & \textbf{Slope} \\
        \midrule

        \multicolumn{5}{l}{\textbf{Ground}}\\
        Pitch           (\unit{\degree}) & $5$   & --    & --    & -- \\
        Roll            (\unit{\degree}) & $5$   & --    & --    & -- \\
        Noise amplitude (\unit{\meter})  & $0.1$ & $0.5$ & $0.5$ & $0.5$ \\
        \addlinespace

        \multicolumn{5}{l}{\textbf{Contacts}}\\
        Friction coefficient & $[0.5,\,0.9]$ & $[0.5,\,0.9]$ & -- & -- \\
        \addlinespace

        \multicolumn{5}{l}{\textbf{Initial rover pose}}\\
        Yaw angle (\unit{\degree}) & $180$ & $45$ & $45$ & $45$ \\
        \addlinespace

        \multicolumn{5}{l}{\textbf{Desired path}}\\
        Lateral offset (\unit{\meter})  & $2$ & $1$ & $1$  & $1$ \\
        Heading offset (\unit{\degree}) & --  & --  & $20$ & $20$ \\
        \addlinespace

        \multicolumn{5}{l}{\textbf{Rocks/Step}}\\
        Density (\unit{\per\meter\squared}) & $0.5$         & --            & -- & -- \\
        Length  (\unit{\meter})             & $[0.2,\,0.4]$ & --            & -- & -- \\
        Width   (\unit{\meter})             & $[0.2,\,0.6]$ & --            & -- & -- \\
        Height  (\unit{\meter})             & $[0.2,\,0.6]$ & $[0.2,\,0.3]$ & -- & -- \\
        Roll    (\unit{\degree})            & $20$          & $1$           & -- & -- \\
        Pitch   (\unit{\degree})            & $20$          & $1$           & -- & -- \\
        Yaw     (\unit{\degree})            & $90$          & $20$          & -- & -- \\
        Position (\unit{\meter})            & --            & $[1.2,\,1.6]$ & -- & -- \\
        \addlinespace

        \multicolumn{5}{l}{\textbf{Ripples geometry}}\\
        Amplitude          (\unit{\meter}) & -- & -- & $[0.1,\,0.3]$ & -- \\
        Wavelength         (\unit{\meter}) & -- & -- & $[0.6,\,1]$   & -- \\
        Smoothing $\alpha$                 & -- & -- & $[0,\,0.2]$   & -- \\
        \addlinespace

        \multicolumn{5}{l}{\textbf{Slope geometry}}\\
        Slope angle     (\unit{\degree})    & -- & -- & -- & $[15,\,30]$ \\
        $\beta_{\mathrm{start}}$ (\unit{\per\meter}) & -- & -- & -- & $[0.1,\,0.5]$ \\
        Length          (\unit{\meter})     & -- & -- & -- & $[1.5,\,3]$ \\
        $\beta_{\mathrm{end}}$   (\unit{\per\meter}) & -- & -- & -- & $[0.2,\,2]$ \\
        \bottomrule

    \end{tabular}
    \caption{Randomization parameters for each scenario. If a range is specified, the value at each new trial is drawn from a uniform distribution within these bounds. Otherwise, a single value indicates a maximum bound. $\beta_{\mathrm{start}}$ and $\beta_{\mathrm{end}}$ are the quadratic coefficients of the parabolic sections of terrain respectively before (concave) and after (convex) the slope.}
    \label{tab:rand_params}
\end{table}

\subsubsection{Rock fields}

The base terrain of the rock-field scenario is globally planar, with irregularities introduced through Perlin noise, and is randomly inclined in both roll and pitch by up to \qty{5}{\degree}. Rigid obstacles (shown in gray in Fig.~\ref{fig:darts_scenarios}) are superimposed on this surface. These obstacles consist of rectangular parallelepipeds whose centers lie on the terrain surface. Wheel--obstacle interactions follow a stiff spring--damper contact model with Coulomb friction, with the friction coefficient randomly sampled between \qty{0.5} and \qty{0.9} at each trial. The obstacles are generated with an average spatial density $\rho$ ranging from \qty{0} to \qty{0.5} elements per square meter. To randomize their spatial distribution, an instance probability $p_{\mathrm{instance}} = 0.1$ is introduced: within each area of size $(p_{\mathrm{instance}})^2/\rho$, an obstacle is placed with probability $p_{\mathrm{instance}}$, and its position is sampled uniformly within that region. The obstacle dimensions and orientations are randomized within the ranges specified in Table~\ref{tab:rand_params}.

In this scenario, the rover is also trained to robustly steer and track the desired path from large initial deviations, with heading errors of up to \qty{180}{\degree} and lateral offsets of up to \qty{2}{\meter}.

\subsubsection{Step Obstacles}

The step obstacle consists of a single rectangular parallelepiped whose height and orientation are randomized, while its width and length are assumed to be effectively infinite. The position of the step relative to the rover’s initial pose is also randomized, ensuring that the obstacle is encountered at varying times within the policy period.

This obstacle is explicitly chosen because it is a canonical yet particularly challenging case, commonly used in the literature to benchmark six-wheeled suspensions, and typically intractable for four-wheeled rovers.

\subsubsection{Ripples}

\begin{figure}
	\centering
	\includegraphics[width=0.4\textwidth]{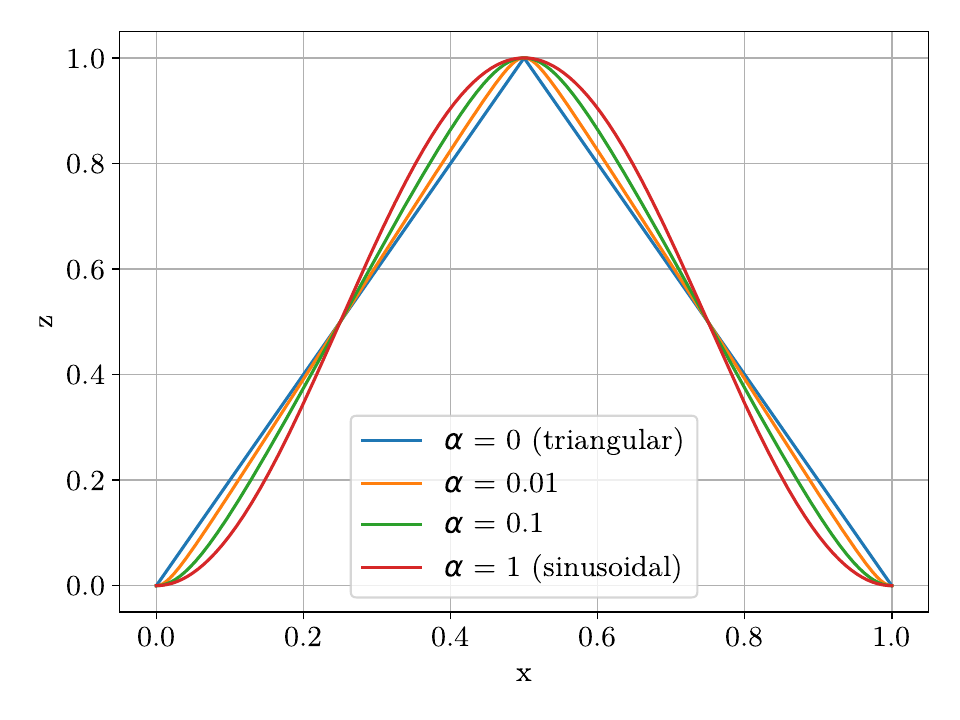}
	\caption{Function used to define the shape of the ripples, for a wavelength and amplitude of 1. $\alpha$ is a smoothing parameter that determines the shape of the ripple, from triangular wave to sinusoidal.}
	\label{fig:ripple_function}
\end{figure}

The shape of the ripples is defined by the following equation:
\begin{equation}
    z(x) = \frac{A}{2} \left(
    \frac{
    \left[ 1 - \frac{2}{\pi} \cos^{-1}\!\left( (1-\alpha)\sin\!\left(\frac{2\pi x}{\lambda}\right) \right) \right]
    }{
    \left[ 1 - \frac{2}{\pi} \cos^{-1}(1-\alpha) \right]
    }
    + 1 \right)
    \label{eq:ripple}
\end{equation}
where $\lambda$ denotes the wavelength and $\alpha$ is a smoothing parameter controlling the ripple shape, ranging from triangular ($\alpha = 0$) to sinusoidal ($\alpha = 1$), as illustrated in Fig.~\ref{fig:ripple_function}. As in the other scenarios, Perlin noise is subsequently added to introduce irregularities.

In this scenario, each terrain instance consists of four consecutive ripples, exposing the rover to the transition from rippled to flat terrain and enabling it to learn to negotiate both the first and final ripples. Each consecutive ripple has distinct geometry, with independently sampled amplitude, wavelength, and smoothing parameter.

The desired path requires the rover to traverse the ripples at a random angle of up to \qty{20}{\degree} relative to the wave direction, while the rover is initialized with a random heading offset of up to \qty{45}{\degree}.

\subsubsection{Slopes}

In this scenario, the rover starts on horizontal ground ahead of a slope that transitions to its nominal incline through a parabolic profile with quadratic coefficient $\beta_{\mathrm{start}}$. After a constant-slope section, the terrain transitions back to horizontal through a second parabolic profile with quadratic coefficient $\beta_{\mathrm{end}}$. If the randomly sampled length, defined as the distance between the start of the first parabolic transition and the end of the second, is too short relative to the independently sampled values of $\beta_{\mathrm{start}}$ and $\beta_{\mathrm{end}}$, the two parabolic profiles connect directly, without an intermediate constant-slope section.

The desired path requires the rover to climb at a random angle of up to \qty{20}{\degree} relative to the slope, while the rover is initialized with a random heading offset of up to \qty{45}{\degree}.

\subsection{POLICY CONSOLIDATION}

Once a proficient policy has been obtained for each of the four terrain classes, the most recent one million samples of experience from each training instance are aggregated, and the neural networks used to generate them are discarded. Each sample consists of a transition tuple composed of two observed states, the action applied between them, and the associated reward, corresponding to a simulated transition between two consecutive decision steps.

A new set of actor and critic networks is then trained exclusively on this aggregated dataset using the same learning algorithm, but with slightly modified hyperparameters, as detailed in Table~\ref{tab:consolidation_parameters}. Because the dataset already captures near-optimal behaviors for each terrain type, substantially fewer training iterations are required to obtain an effective policy. In addition, the minibatch size is increased fourfold to promote smoothness and consistency of the learned policy across all terrain conditions. The hyperparameters that remain the same across both training phases are listed in Table~\ref{tab:rl_parameters}.

During the initial training phase, six simulation instances are run in parallel for each scenario, resulting in a total of 24 independent DARTS simulations that continuously collect experience while the policy is updated between trials. This process requires approximately 24 hours on a high-performance computing cluster to obtain a satisfactory policy for every scenario. In contrast, the consolidation phase does not involve further simulation and requires fewer training iterations, reducing its duration to approximately 15 minutes.

Although directly training a single policy across all scenarios simultaneously proved intractable, the actor network obtained from the aggregated experience yields a unified controller that alone can handle all four terrain types.

\begin{table}
    \centering
    \begin{tabular}{lcc}
        \toprule
        \textbf{Training parameter} & \textbf{Independent scenarios} & \textbf{Consolidation} \\
        Replay buffer size    & \num{1e6}  & \num{4e6}  \\
        Minibatch size        & 128        & 512        \\
        Initial learning rate & \num{1e-5} & \num{1e-3} \\
        Number of iterations  & \num{5e6}  & \num{1e4}  \\
        \bottomrule
    \end{tabular}
    \vspace{1.5ex}
    \caption{Comparison of hyperparameters used for training in independent scenarios and for the subsequent policy consolidation phase.}
    \label{tab:consolidation_parameters}
\end{table}

\begin{table}
    \centering
    \begin{tabular}{lccc}
        \toprule
        \textbf{Training parameter}        & \textbf{Symbol} & \textbf{Value}  \\
        State dimension                    & $n_s$           & 45              \\
        Action dimension                   & $n_a$           & 3               \\
        Policy evaluation rate             & $f_\pi$         & \qty{2}{\hertz} \\
        Reward discount factor             & $\gamma$        & 0.99            \\
        Soft target update factor          & $\tau$          & \num{5e-3}      \\
        Policy update delay                & $d$             & 2               \\
        Target policy regularization noise & $\epsilon$ & $\mathcal{N}(0,\, 0.05^2)$ \\
        Bounds of the target policy regularization & $c$ & 0.2 \\
        \bottomrule
    \end{tabular}
    \vspace{1.5ex}
    \caption{Hyperparameters used in both training phases.}
    \label{tab:rl_parameters}
\end{table}

\section{RESULTS}
\label{sec:results}

The resulting neural network controller is directly deployed on the physical rover and evaluated on an obstacle course comprising all terrain types represented in the training scenarios. The same controller operates continuously throughout the experiment and is responsible for maneuvering the rover along the desired path. This path is defined as a sequence of line segments automatically provided to the controller based on a predefined route and the rover’s pose estimated from visual--inertial odometry. The rover has no prior knowledge of the terrain to be traversed. Instead, the local terrain elevation ahead is reconstructed in real time using stereo vision, while the force–torque sensors at each wheel assembly provide feedback on wheel--ground interactions. All computations are performed onboard the rover.

\subsection{ROCK FIELD}

\begin{figure}
	\centering
	\newcommand{\widthratio}{0.24}

	\renewcommand{\thesubfigure}{a}%
	\subfloat[]{%
        \includegraphics[width=\widthratio\textwidth]{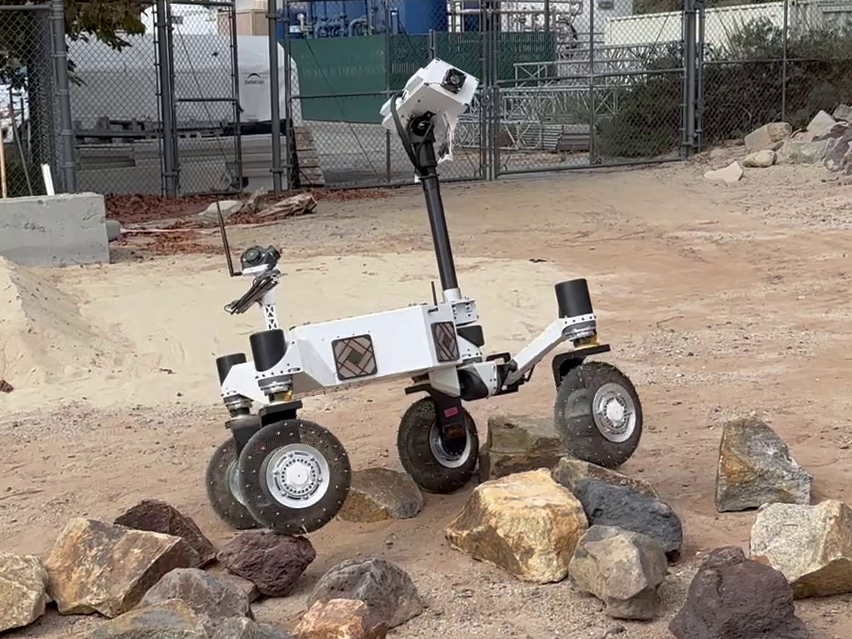}%
        \label{fig:rock_rl_1}%
    }
    \hfil
    \renewcommand{\thesubfigure}{b}%
	\subfloat[]{%
        \includegraphics[width=\widthratio\textwidth]{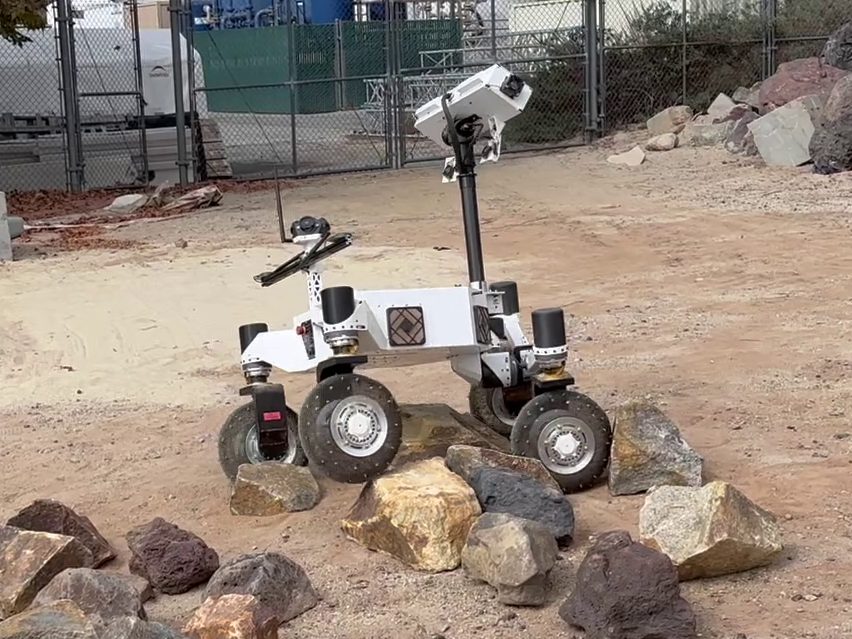}%
        \label{fig:rock_rl_2}%
    }

	\caption{The resulting neural-network controller tested over a rock field.}
	\label{fig:rock_rl}
\end{figure}

Fig.~\ref{fig:rock_rl} shows the rover traversing a rock field. We observe that the yaw joint of the Active Gimbal Suspension rotates left and right in response to incoming rocks, helping to sequence the climbing of the wheels one at a time. This is indeed the most effective strategy for a four-wheeled vehicle, which can lift only one wheel at a time. Meanwhile, the roll joint continuously adjusts the bogie angle to conform to the geometry of the rocks, and no wheel is left suspended. This controlled rotation of the bogie assists the wheels in climbing over obstacles by actively shifting the load to the adjacent wheels. We also observe that the controller remains robust to rocks rolling or shifting beneath the rover, owing to the direct force feedback from each wheel assembly.

\subsection{BICKLER TRAP}

We arranged the rocks into a continuous row that can be straddled by the rover. This type of obstacle is commonly referred to as a \emph{bump}, or a \emph{Bickler trap}, named after the inventor of the rocker–bogie suspension, who identified the particular difficulty such configurations pose for passive suspension systems. Because the obstacle can be straddled, once the front wheels have cleared it, they return to the same elevation as the rear wheels before the latter begin climbing. As a result, the pull generated by the front wheels opposes the ascent of the rear wheels when the obstacle is sufficiently steep. Consequently, the rear wheels fail to surmount the obstacle, and the symmetry of the configuration prevents the front wheels from re-clearing it when reversing, thereby immobilizing the vehicle.

\begin{figure}
	\centering
	\newcommand{\widthratio}{0.24}

	\renewcommand{\thesubfigure}{a}%
	\subfloat[]{%
        \includegraphics[width=\widthratio\textwidth]{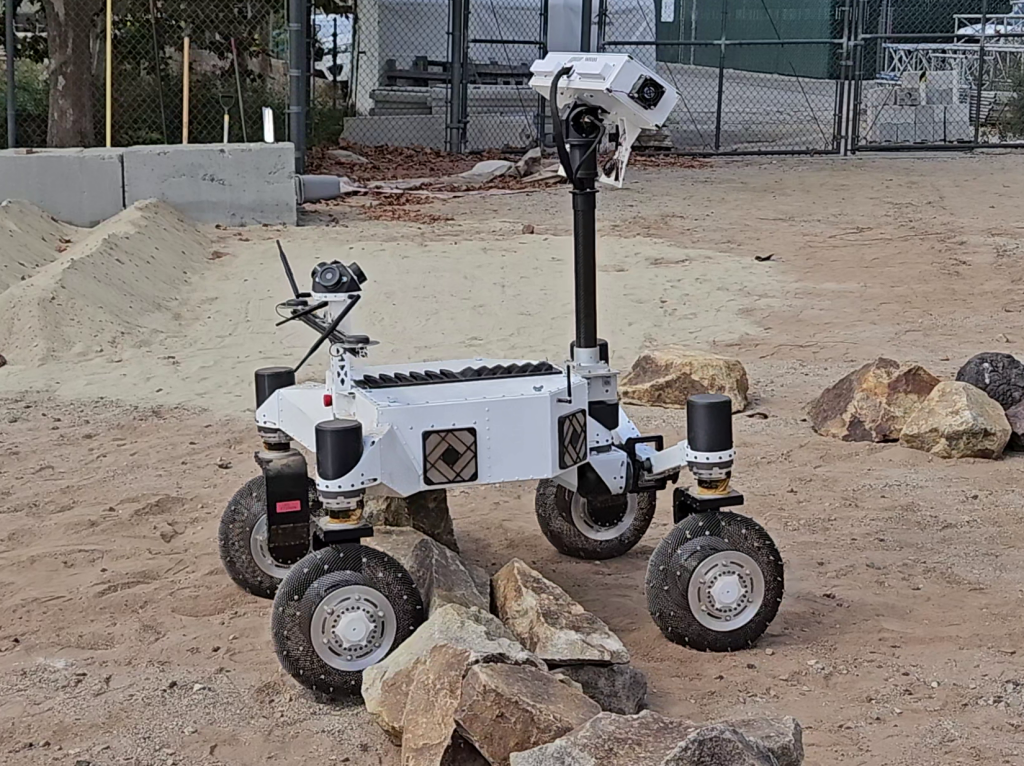}%
        \label{fig:bickler_passive_1}%
    }
    \hfil
    \renewcommand{\thesubfigure}{b}%
	\subfloat[]{%
        \includegraphics[width=\widthratio\textwidth]{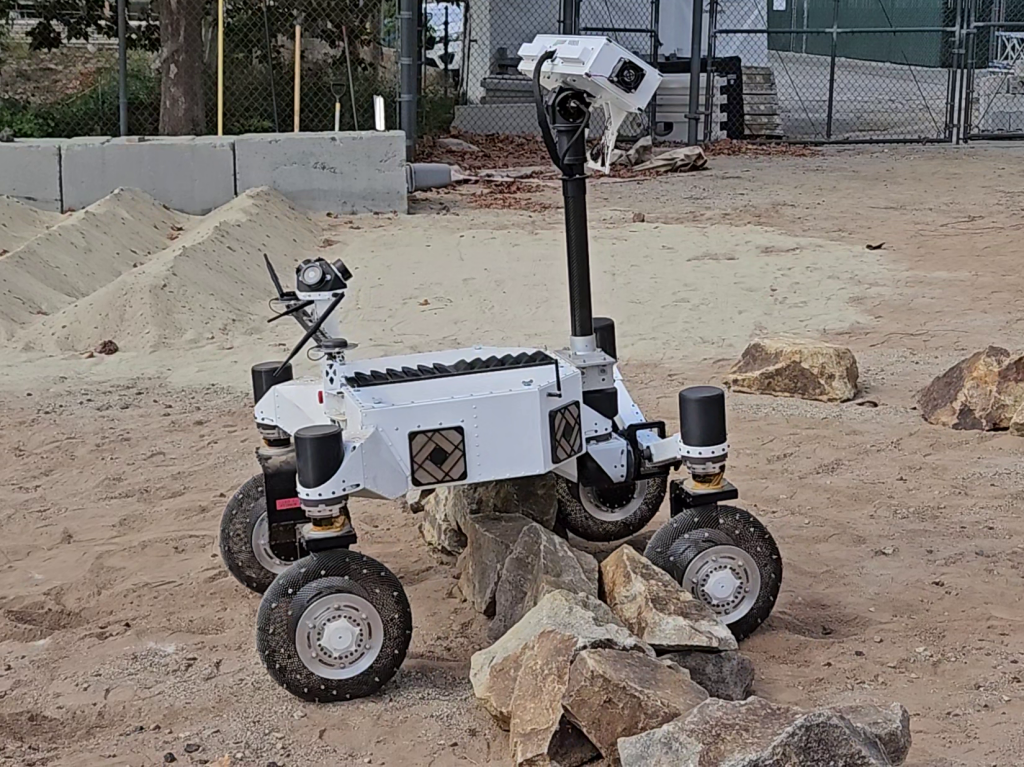}%
        \label{fig:bickler_passive_2}%
    }

	\caption{Passive bogie suspension driving over a Bickler trap. \protect\subref{fig:bickler_passive_1} The rear wheels cannot overcome the obstacle while driving forwards. \protect\subref{fig:bickler_passive_2} Similarly, the front wheels cannot clear the obstacle again when driving backwards, trapping the vehicle.}
	\label{fig:bickler_passive}
\end{figure}

Fig.~\ref{fig:bickler_passive} shows the rover attempting to traverse such a bump with the Active Gimbal Suspension disabled and the roll-joint clutch open, effectively resulting in a free bogie suspension. As expected, the front wheels successfully pass the obstacle, whereas the rear wheels do not. Furthermore, when reversing, the front wheels are unable to climb the bump again, leaving the rover trapped.

\begin{figure}
	\centering
	\newcommand{\widthratio}{0.24}

	\renewcommand{\thesubfigure}{a}%
	\subfloat[]{%
        \includegraphics[width=\widthratio\textwidth]{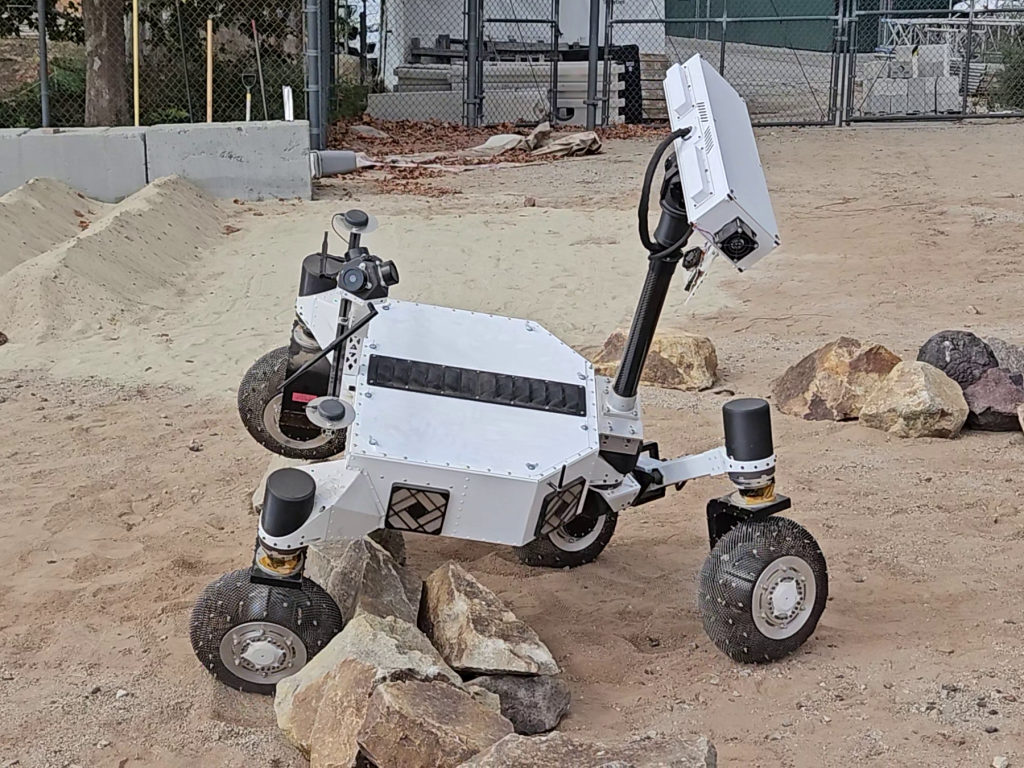}%
        \label{fig:bickler_rl_1}%
    }
    \hfil
    \renewcommand{\thesubfigure}{b}%
	\subfloat[]{%
        \includegraphics[width=\widthratio\textwidth]{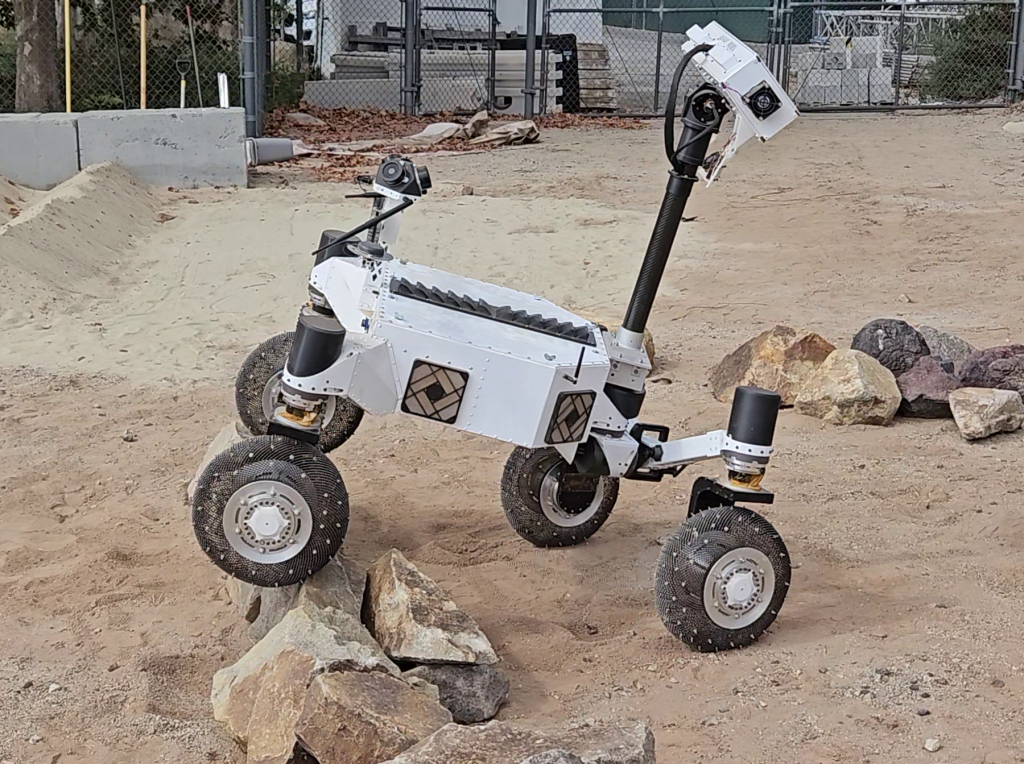}%
        \label{fig:bickler_rl_2}%
    }

	\caption{The neural-network controller handling the Bickler trap.}
	\label{fig:bickler_rl}
\end{figure}

Fig.~\ref{fig:bickler_rl} shows the rover traversing the same bump using the Active Gimbal Suspension and the neural network controller. In this case, the yaw and roll joints actively reconfigure the rover to position one of the rear wheels atop the obstacle. The roll joint then reverses its motion to assist in lifting and clearing the final wheel. Although this obstacle configuration was not explicitly represented in the training scenarios, the learned policy successfully handles it, demonstrating effective generalization.

\subsection{STEP}

\begin{figure}
	\centering
	\newcommand{\widthratio}{0.24}

	\renewcommand{\thesubfigure}{a}%
	\subfloat[]{%
        \includegraphics[width=\widthratio\textwidth]{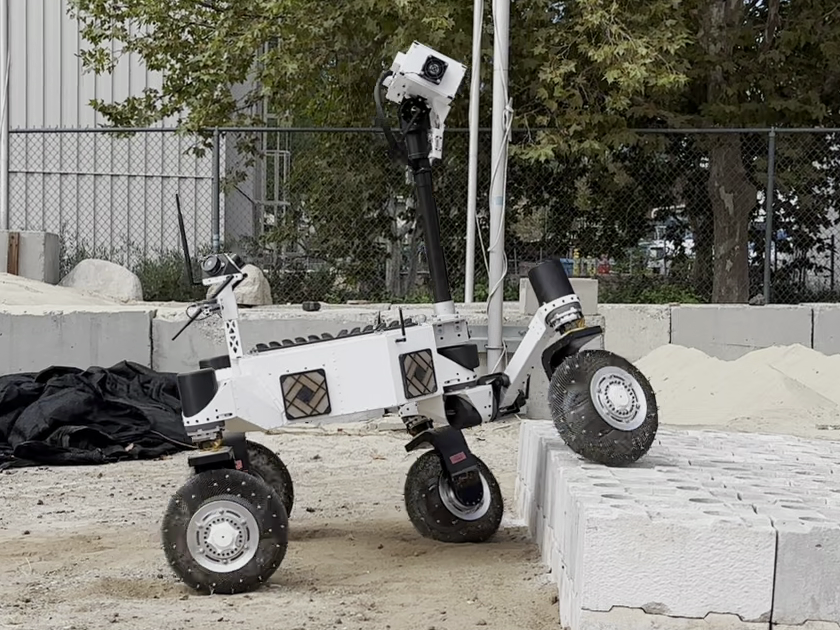}%
        \label{fig:step_rl_1}%
    }
    \hfil
    \renewcommand{\thesubfigure}{b}%
	\subfloat[]{%
        \includegraphics[width=\widthratio\textwidth]{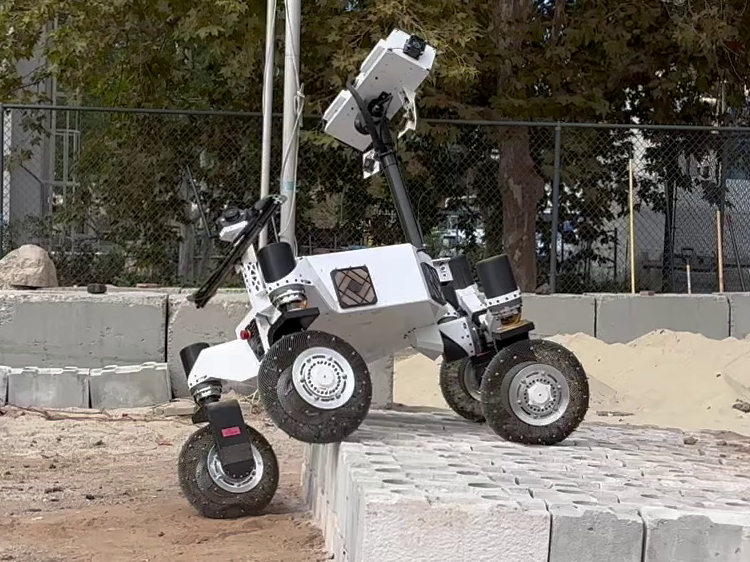}%
        \label{fig:step_rl_2}%
    }

	\caption{The neural-network controller climbing over a wheel-high step.}
	\label{fig:step_rl}
\end{figure}

\begin{figure}
	\centering
	\includegraphics[width=0.5\textwidth]{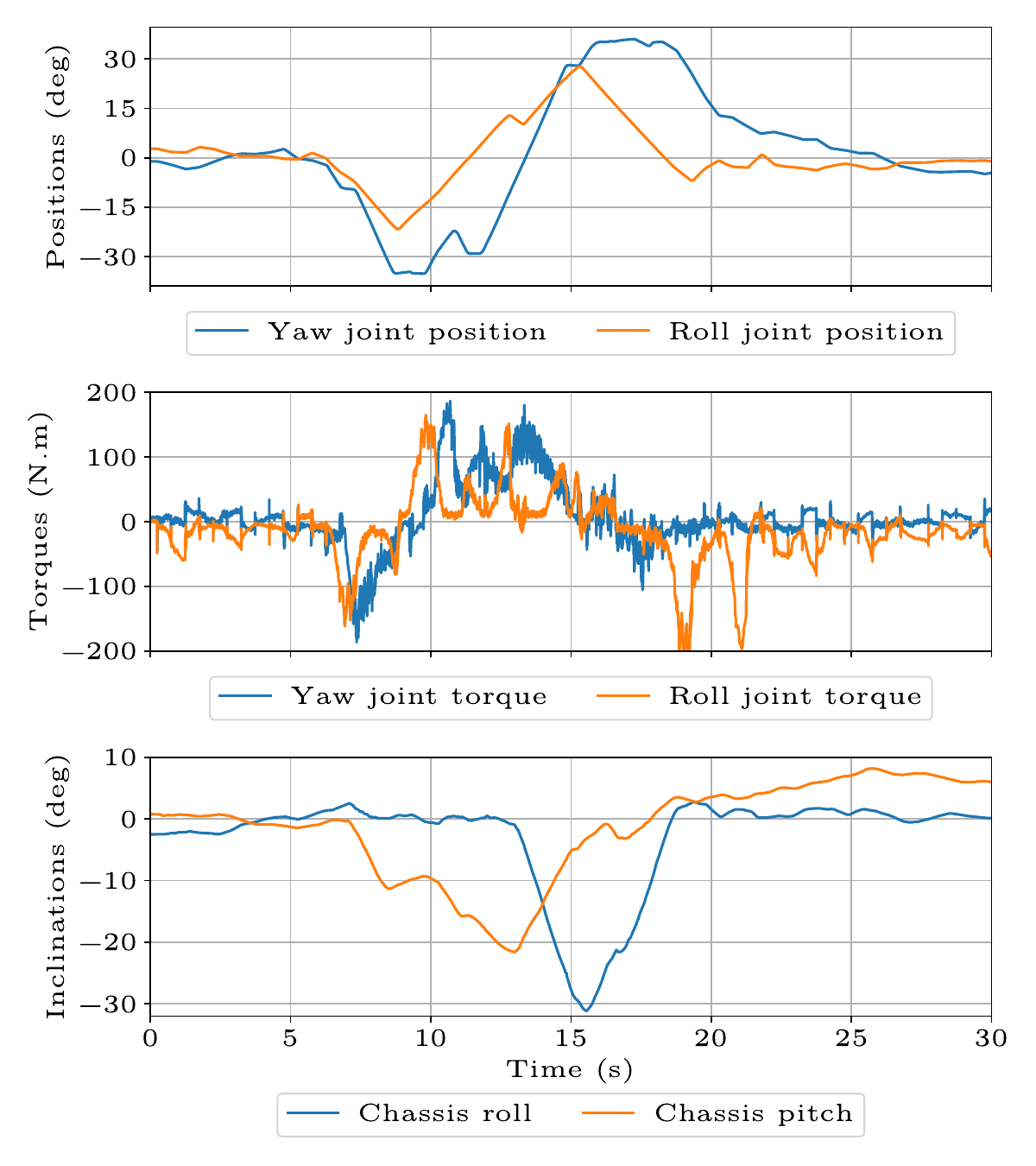}
	\caption{Action of the Active Gimbal Suspension while the rover is climbing over the step obstacle. The torque in the joints is estimated from the winding current, the torque constant, and the gearbox efficiency of the actuator. The rotation axis of the yaw joint is pointing downward and the roll-joint axis forward, therefore the positions and torques are negative when moving the right-front wheel backward and downward relative to the chassis.}
	\label{fig:exp_step_climb}
\end{figure}

To enable the rover to climb a step obstacle, the controller actuates both joints of the Active Gimbal Suspension such that the wheels clear the obstacle sequentially, as shown in Fig.~\ref{fig:step_rl}. Isolating each wheel climb allows the roll joint to provide targeted lift assistance by alternating the torque applied to the bogie. Fig.~\ref{fig:exp_step_climb} shows negative roll-joint torque assisting the right-front and left-rear wheels, and positive torque assisting the other two wheels as they climb in succession.

\subsection{RIPPLES}

When operating with a passive suspension, the wheels can rapidly become embedded and entrapped in sand ripples whose wavelength is comparable to the vehicle wheelbase. In this configuration, all wheels simultaneously face an uphill slope, despite the chassis remaining approximately horizontal.

The ripples considered in this study are composed of M90, a Mars soil simulant analogous to the material found in Martian dunes and ripples~\cite{oravec2021geotechnical}. They have a crest-to-crest spacing of \qty{80}{\cm} and are formed up to the angle of repose, approximately \qty{35}{\degree}, resulting in a height of about \qty{28}{\cm}.

To traverse this terrain, the controller operates the Active Gimbal Suspension to lift and position one of the front wheels beyond the ripple crest. This wheel is then used as an anchor point: the rover pulls itself forward by shifting the load onto the corresponding bogie arm and rotating it backward, thereby propelling the rover and the remaining wheels over the crests.

On ripples with periodic and uniform geometry, this behavior may resemble a gait. However, the controller continuously adapts its motion based on the local terrain geometry, the desired path, and the force–torque feedback from each wheel assembly. For instance, if the anchor wheel does not extend sufficiently beyond the crest, the controller may attempt to advance it further before initiating the pull, resulting in a rapid back-and-forth crawling motion, or alternatively reverse the bogie motion completely to engage the opposite wheel as the anchor.

\begin{figure}
	\centering
	\newcommand{\widthratio}{0.24}

	\renewcommand{\thesubfigure}{a}%
	\subfloat[]{%
        \includegraphics[width=\widthratio\textwidth]{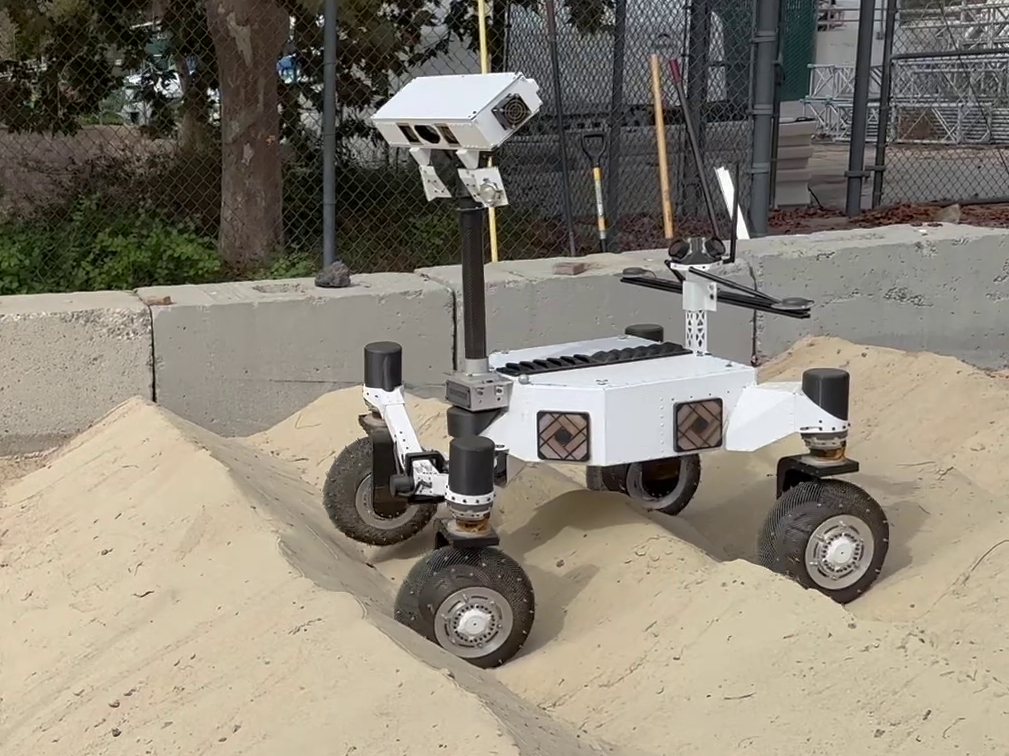}%
        \label{fig:ripples_rl_1}%
    }
    \hfil
    \renewcommand{\thesubfigure}{b}%
	\subfloat[]{%
        \includegraphics[width=\widthratio\textwidth]{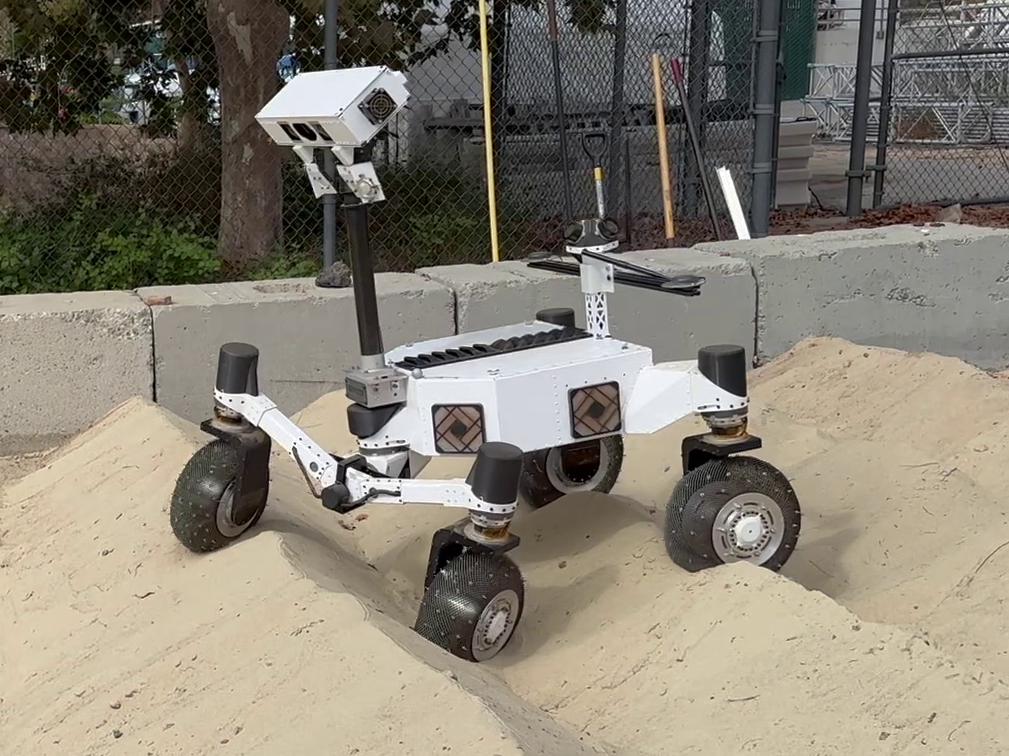}%
        \label{fig:ripples_rl_2}%
    }

	\caption{The neural-network controller traversing sand ripples.}
	\label{fig:ripples_rl}
\end{figure}

\subsection{SLOPE}

\begin{figure}
	\centering
	\newcommand{\widthratio}{0.24}

	\renewcommand{\thesubfigure}{a}%
	\subfloat[]{%
        \includegraphics[width=\widthratio\textwidth]{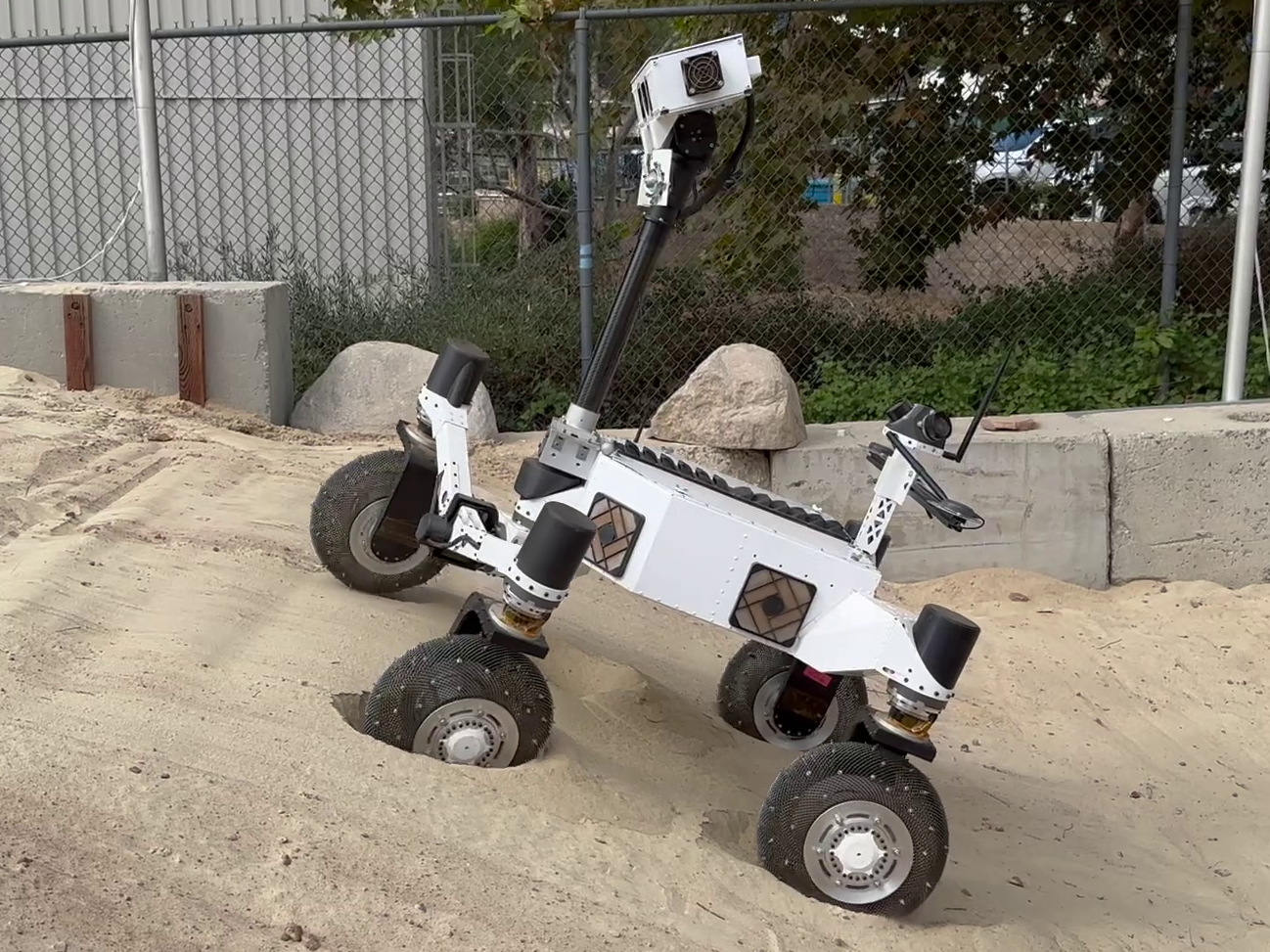}%
        \label{fig:slope_rl_1}%
    }
    \hfil
    \renewcommand{\thesubfigure}{b}%
	\subfloat[]{%
        \includegraphics[width=\widthratio\textwidth]{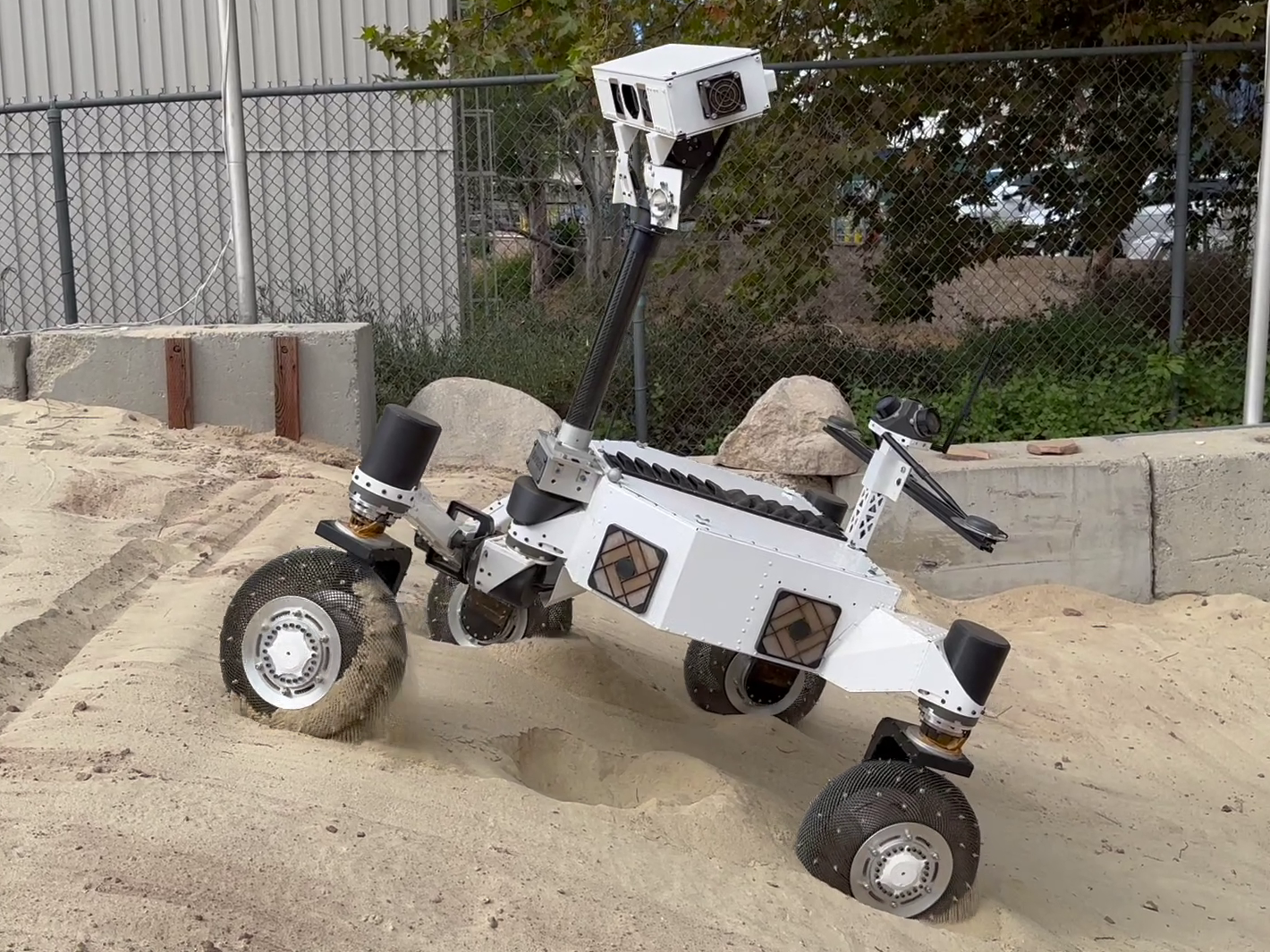}%
        \label{fig:slope_rl_2}%
    }

	\caption{The neural-network controller climbing a \qty{20}{\degree} slope.}
	\label{fig:slope_rl}
\end{figure}

Fig.~\ref{fig:slope_rl} shows the rover climbing a \qty{20}{\degree} slope composed of M90 Mars soil simulant. To mitigate wheel sinkage and enable efficient ascent, the rover adopts a crawling gait in which the load is alternately transferred to the left and right wheels to serve as anchors using the roll joint, while the yaw joint rotation pulls the chassis forward. This gait was first proposed in~\cite{bouton2022crawling}, and its spontaneous emergence as the optimal policy on soft slopes under the Bekker--Wong terramechanics model provides a meaningful validation.

\begin{figure}
	\centering
	\includegraphics[width=0.5\textwidth]{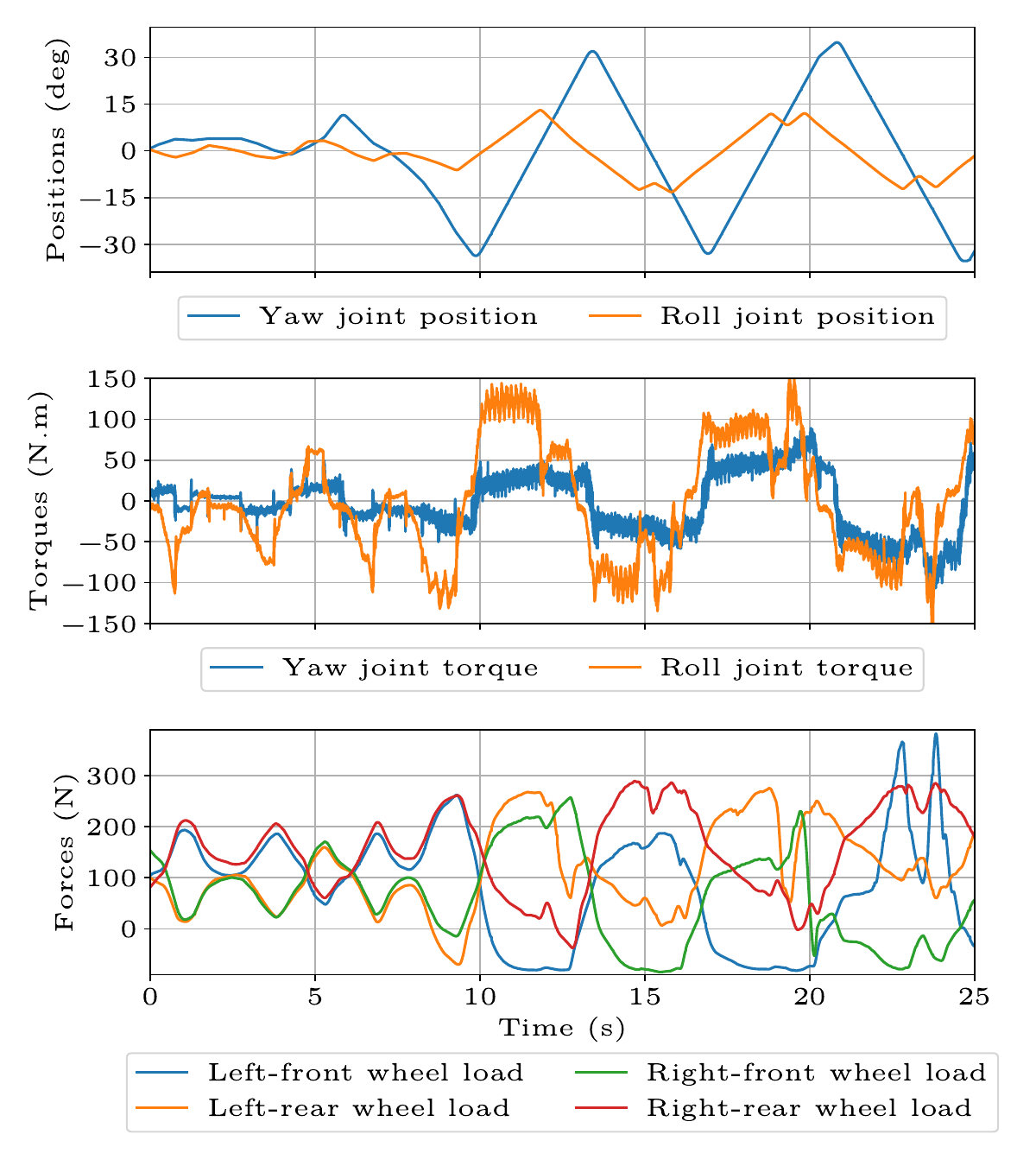}
	\caption{Resulting gait when the physical rover climbs a \qty{20}{\degree} slope. The torque in the gimbal joints are estimated from the winding current, the torque constant, and the gearbox efficiency of the actuator.}
	\label{fig:exp_slope_gait}
\end{figure}

\begin{figure}
	\centering
	\includegraphics[width=0.5\textwidth]{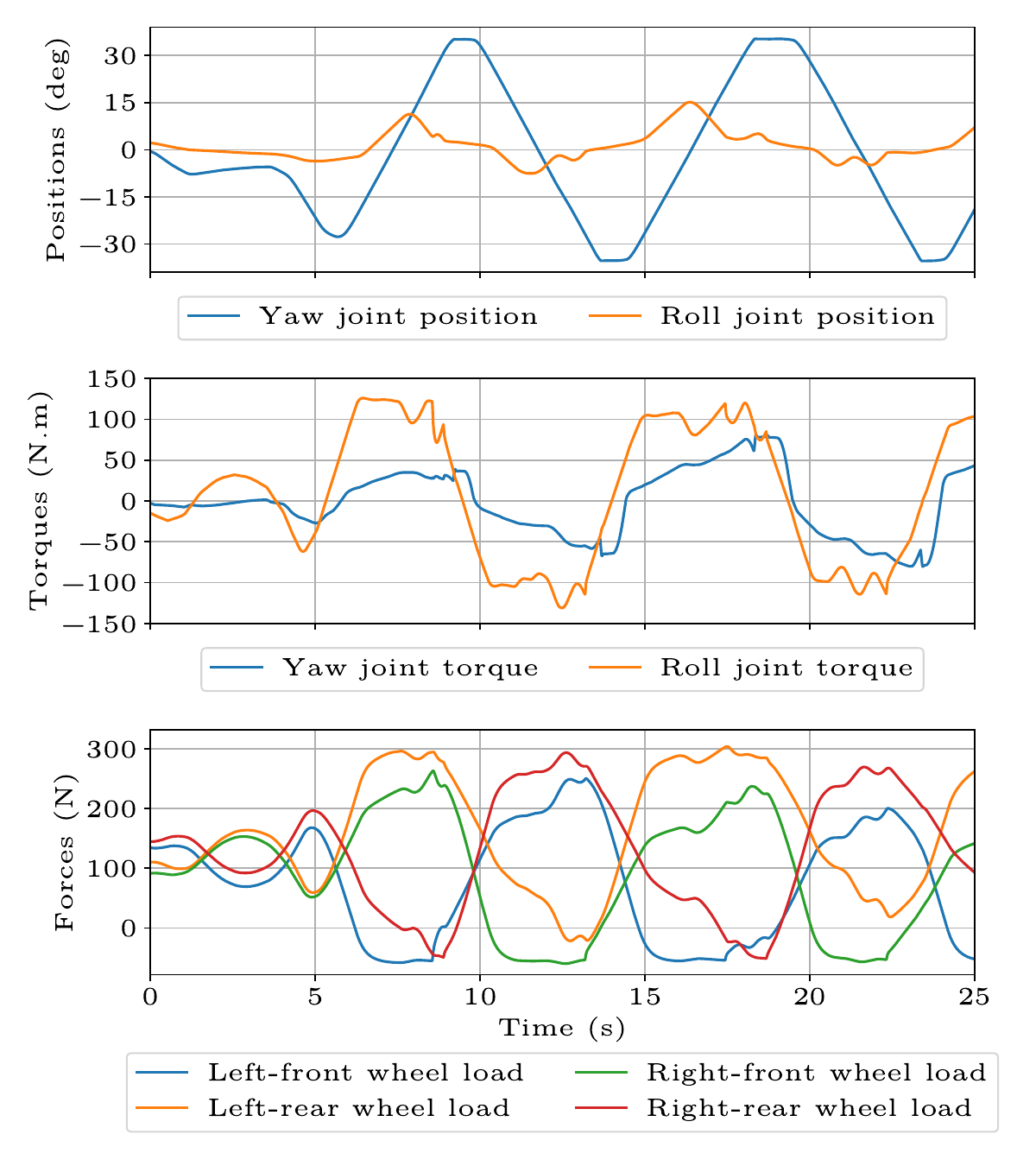}
	\caption{Resulting gait on a \qty{20}{\degree} slope in simulation. The vertical forces are evaluated above the wheels, at the locations corresponding to the force–torque sensors on the physical rover, hence the values becoming negative when a wheel is fully unloaded, due to the weight of the wheel assembly itself.}
	\label{fig:sim_slope_gait}
\end{figure}

The action of the Active Gimbal Suspension during this gait, as executed on the physical rover, is shown in Fig.~\ref{fig:exp_slope_gait}, and can be compared to the gait obtained in simulation on a similar slope using the terramechanics modeling in Fig.~\ref{fig:sim_slope_gait}. In both cases, the controller provides a smooth transition from flat ground to the slope. However, the roll-joint position does not closely follow the same trajectory, highlighting the adaptive nature of the controller and the influence of real-time feedback from the actual terrain geometry and wheel--ground interaction forces.

\begin{figure}
	\centering
	\includegraphics[width=0.5\textwidth]{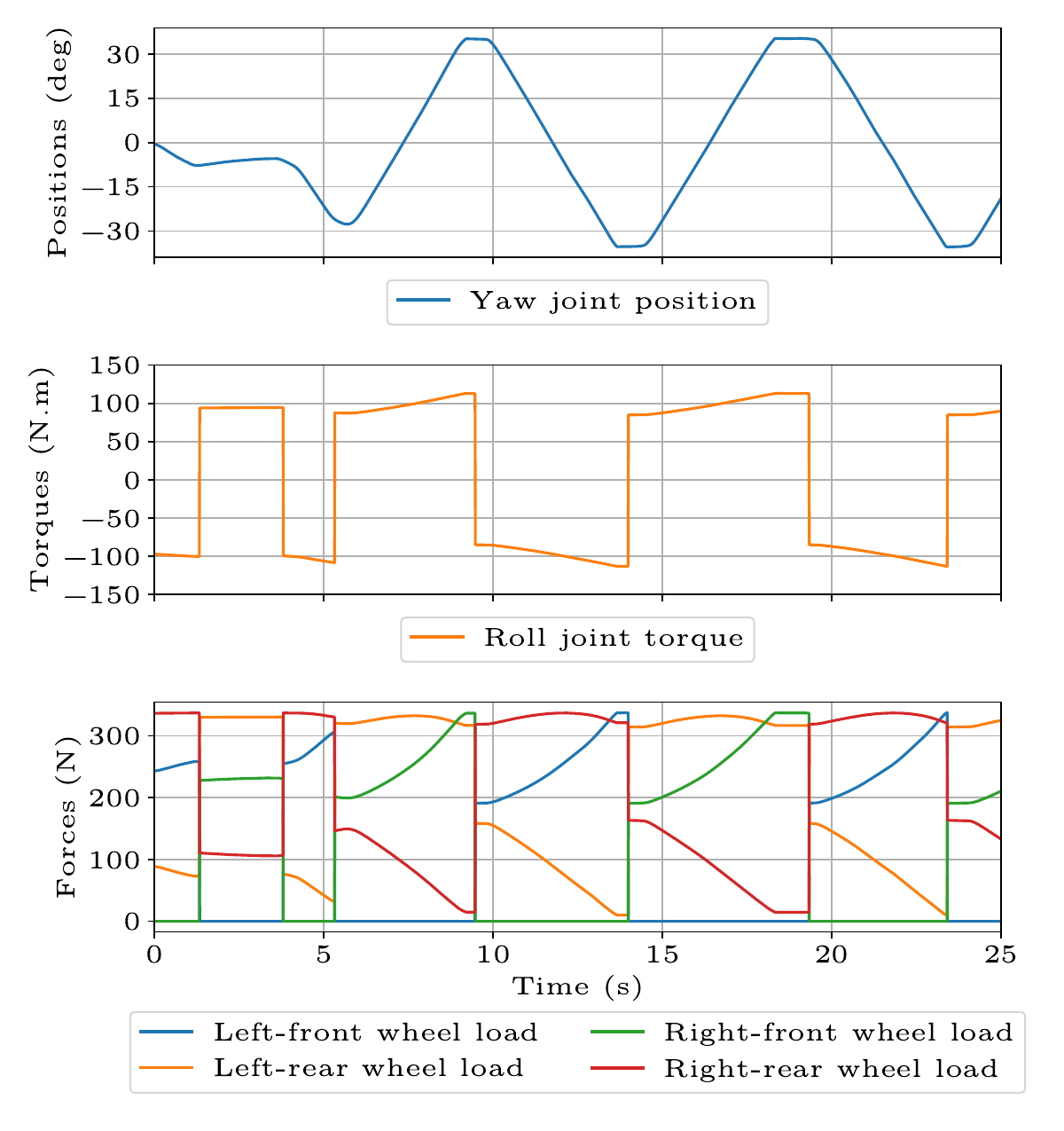}
	\caption{Maximum torque and load shift that can theoretically be applied on a \qty{20}{\degree} slope across the different configurations according to the mass model. The loads are directly the vertical forces between the wheels and the ground.}
	\label{fig:slope_load}
\end{figure}

The torque applied to the bogie by the controller during the slope ascent can be compared to the theoretical maximum torque that can be applied before inducing wheel lift-off. This limiting torque, shown in Fig.~\ref{fig:slope_load}, along with the resulting load distribution across the wheels, is computed via linear programming for each configuration on a \qty{20}{\degree} slope using the mass model of ERNEST derived in Section~\ref{sec:mass_model}. We observe that the controller operates close to this theoretical limit, effectively maximizing the unloading of the advancing wheel.

Table~\ref{tab:slope_perf} presents a performance comparison with a passive suspension. The travel reduction is defined as the difference between the expected and actual travel distances over a given time window, normalized by the expected travel distance. The expected travel distance is obtained by integrating the rover motion under a no-slip assumption. For the learned controller, which exhibits a cyclic gait, the time window considered spans a full gait cycle. The cost of transport is computed from the cumulative electrical energy consumption of all actuators, normalized by the actual travel distance and the rover mass. The electrical energy is obtained by integrating the power, estimated as
\begin{equation}
P = I \left( R I + K_{\tau} ,\omega \right),
\end{equation}
where $I$ denotes the winding current, $R$ the winding resistance, $K_{\tau}$ the torque constant, and $\omega$ the angular velocity. Despite the additional energy consumption introduced by the two actuators of the Active Gimbal Suspension, the improved motion efficiency achieved by the neural network controller on sandy slopes results in a lower overall cost of transport than a passive suspension relying solely on wheel drives.

This performance gap further increases on wet sand, as reported in Table~\ref{tab:slope_perf}. The stick--slip interaction between the wheels and the wet sand induces significant vibrations, which prevent the passive suspension from making forward progress, ultimately leading to complete entrapment of the rover. In contrast, the increased cohesion of the wet sand enhances the effectiveness of the crawling gait, thereby also demonstrating the controller’s robustness to soil properties outside the training domain.

\begin{table}
    \centering
    \setlength{\tabcolsep}{4pt}
    \begin{tabular}{lccc}
        \hline
        \textbf{Performance index} & \textbf{Sand} & \textbf{Passive bogie} & \textbf{RL controller} \\
        \hline
        \multirow{2}{*}{Travel reduction (\unit{\percent})} & Dry & 81.9 & 43.0 \\
        & Wet & 100 & 30.0 \\
        \hline
        \multirow{2}{*}{Cost of transport (\unit{\milli\watt\hour\per\meter\per\kilogram})} & Dry & 21.3 & 13.4 \\
        & Wet & $\infty$ & 12.8 \\
        \hline
    \end{tabular}
    \caption{Performance comparison on a \qty{20}{\degree} slope. For the passive bogie configuration, the Active Gimbal Suspension is off and the clutch is disengaged so that the bogie is free to rotate.}
    \label{tab:slope_perf}
\end{table}

\begin{figure}
	\centering
	\newcommand{\widthratio}{0.24}

	\renewcommand{\thesubfigure}{a}%
	\subfloat[]{%
        \includegraphics[width=\widthratio\textwidth]{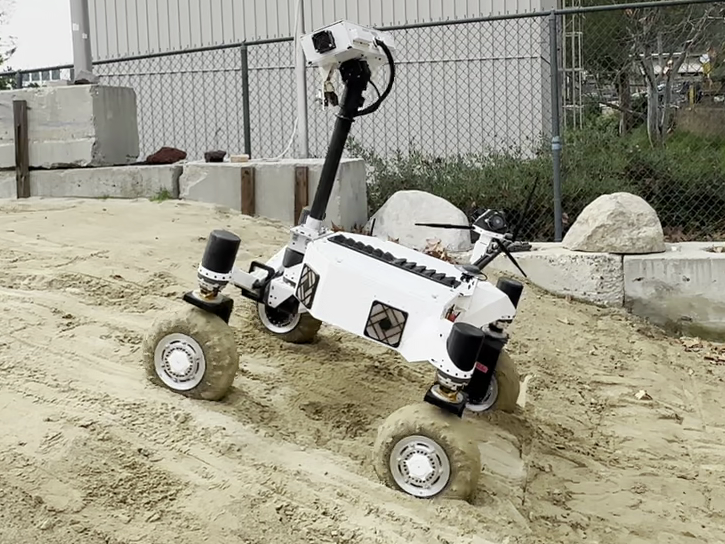}%
        \label{fig:wet_slope_passive}%
    }
    \hfil
    \renewcommand{\thesubfigure}{b}%
	\subfloat[]{%
        \includegraphics[width=\widthratio\textwidth]{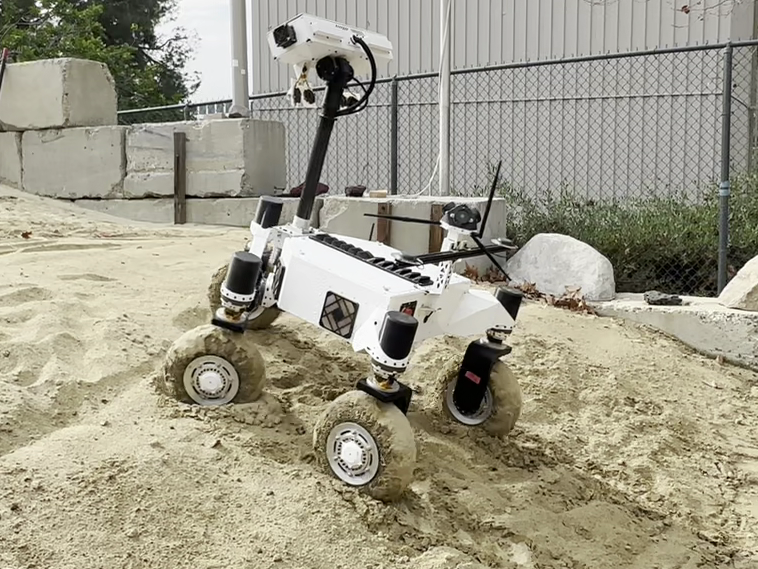}%
        \label{fig:wet_slope_rl}%
    }

	\caption{The rover climbing a \qty{20}{\degree} slope with wet sand. \protect\subref{fig:wet_slope_passive} The passive bogie rapidly becomes fully immobilized. \protect\subref{fig:wet_slope_rl} The RL controller prevents wheel embedment and enables successful ascent of the slope.}
	\label{fig:wet_slope}
\end{figure}

\section{CONCLUSIONS}

This paper presented ERNEST, a four-wheeled planetary rover equipped with a two-degree-of-freedom Active Gimbal Suspension and governed by a neural-network controller trained via reinforcement learning in simulation. By combining steering authority, active load transfer, and obstacle-oriented reconfiguration for sequential wheel placement, the Active Gimbal Suspension enables behaviors that approach some of the versatility of legged locomotion while retaining the simplicity and efficiency of wheeled systems. Experimental results demonstrate that ERNEST can autonomously negotiate rock fields, bumps, wheel-high steps, sand ripples, and loose slopes. Taken together, these findings indicate that a four-wheeled rover with six actuators can surpass the locomotion capabilities of a six-wheeled passive suspension with ten actuators, provided it is paired with an appropriate controller. This is significant because a four-wheeled architecture offers additional advantages: for a given footprint, it can accommodate larger wheels, reducing susceptibility to sinkage and entrapment in depressions, while enabling a simpler and lighter suspension system.

A zero-shot transfer from simulation to the physical rover was achieved through a combination of domain randomization, replay-time sensor noise injection, and a policy consolidation strategy that merges experience collected by terrain-specialized policies into a single unified neural network. This consolidation ensures smooth transitions across heterogeneous environments instead of relying on explicit terrain classification and controller switching. The neural network controller operates on a compact state representation that combines exteroceptive and proprioceptive feedback, including path-tracking variables, sparse terrain elevation samples ahead of each wheel, chassis attitude, joint states, and processed force–torque measurements. The resulting policy exhibits physically relevant behaviors that emerge spontaneously, including sequential wheel climbing with coordinated load redistribution and a crawling gait that independently rediscovers a strategy previously proposed analytically. The policy also demonstrates meaningful generalization to situations not explicitly represented during training, including the Bickler trap and soil conditions whose properties depart from those assumed in the terramechanics model.

Building on these results, future work will extend the approach to incorporate higher-level planning. Domain randomization will first be broadened to cover a wider range of obstacle geometries and their combinations, in order to reduce residual performance gaps on configurations not yet encountered during training. The learned controller will then be combined with a state lattice planner whose heuristic is trained in a supervised manner to reflect the rover’s true mobility capabilities when planning a route toward a target location. Finally, the approach will be evaluated in larger-scale field campaigns conducted in Martian-analog environments.

Beyond these immediate extensions, several broader questions deserve further study. First, it is worth investigating whether incorporating higher-fidelity terramechanics models capturing soil transport and terrain deformation into a subset of the training data could complement and refine the behaviors learned under the lower-fidelity Bekker--Wong formulation. Second, improving robustness to perceptually ambiguous terrain, such as discriminating between rigid and soft slopes, may benefit from exploiting temporal sequences of proprioceptive measurements rather than instantaneous observations. More fundamentally, the use of neural-network controllers on planetary rovers also raises the question of formal validation for safety-critical deployment, which remains an important open challenge for future missions.

\bibliographystyle{IEEEtran}
\bibliography{IEEEabrv,biblio}

\vfill\pagebreak

\end{document}